\pdfoutput=1
\documentclass[11pt]{article}

\usepackage[preprint]{acl}

\usepackage{times}
\usepackage{latexsym}

\usepackage[T1]{fontenc}
\usepackage[utf8]{inputenc}

\usepackage{microtype}

\usepackage{inconsolata}

\usepackage{graphicx}
\usepackage{booktabs}
\usepackage{array}
\usepackage{amssymb}
\usepackage{amsmath} 
\usepackage{hyperref}
\usepackage{multirow} 
\usepackage{makecell}
\usepackage{xcolor}
\usepackage{colortbl}

\usepackage{booktabs}

\newcommand\blfootnote[1]{%
  \begingroup
  \renewcommand\thefootnote{}\footnote{#1}%
  \addtocounter{footnote}{-1}%
  \endgroup
}

\usepackage{xcolor,pifont}
\newcommand*\colourcheck[1]{%
  \expandafter\newcommand\csname #1check\endcsname{\textcolor{#1}{\ding{52}}}%
}
\newcommand*\colourcross[1]{%
  \expandafter\newcommand\csname #1cross\endcsname{\textcolor{#1}{\ding{56}}}%
}
\colourcheck{green}
\colourcheck{orange}
\colourcross{red}

\newcommand{\thedata}{\textit{TransWebEdu}}
\newcommand{\themodel}{\textit{TransWebLLM}}
\newcommand{\themodelweb}{\textit{TransWebLLM-web}}

\newcommand{\themodelcool}{\textit{TransWebLLM-cool}}

\newcommand{\cuatrollm}{\textit{CuatroLLM}}

\usepackage{graphicx}

\title{%
    Multilingual Language Model Pretraining using\\ Machine-translated Data%
}

\author{Jiayi Wang$^{\alpha}$$^\ast$ \quad
Yao Lu$^{\alpha}$$^\ast$ \quad
Maurice Weber$^{\beta}$ \quad
Max Ryabinin$^{\beta}$ \quad
David Adelani$^{\gamma}$ \\
\textbf{Yihong Chen}$^{\alpha}$ \quad
\textbf{Raphael Tang}  \quad
\textbf{Pontus Stenetorp}$^{\alpha, \delta}$ \\
$^{\alpha}$Centre for Artificial Intelligence, University College London \quad \\
$^{\beta}$Together AI \quad $^{\gamma}$Mila, McGill University, Canada CIFAR AI Chair \quad \\
$^{\delta}$Research and Development Center for Large Language Models, National Institute of Informatics\\
\texttt{\{jiaywang,yao.lu,yihong.chen,p.stenetorp\}@cs.ucl.ac.uk}, \\
\texttt{\{maurice,mryab\}@together.ai}, \texttt{david.adelani@mila.quebec} \\
}

\begin{document}
\maketitle
\blfootnote{$^\ast$ Both authors contribute equally to this work.}

\begin{abstract}
High-resource languages such as English, enables the pretraining of high-quality large language models~(LLMs). 
% from community efforts that provide large amounts of high-quality pretraining data.
The same can not be said for most other languages 
as LLMs still underperform for non-English languages, 
% as leading LLMs still show significant performance gaps for non-English languages,
likely due to a gap in the quality and diversity of the available multilingual pretraining corpora.
In this work, we find that machine-translated texts from a single high-quality source language can contribute significantly to the pretraining quality of multilingual LLMs.
We translate \textit{FineWeb-Edu}, a high-quality English web dataset, into nine languages, resulting in a $1.7$-trillion-token dataset, which we call \thedata{} and pretrain a $1.3$B-parameter model, \themodel{}, from scratch on this dataset.
Across nine non-English reasoning tasks, we show that \themodel{} matches or outperforms state-of-the-art multilingual models trained using closed data, such as \textit{Llama3.2}, \textit{Qwen2.5}, and \textit{Gemma}, despite using an order of magnitude less data.
%, such as about 18\% of the tokens used for \textit{Llama3.2}'s training.
%
We demonstrate that adding less than 5\% of \thedata{} as domain-specific pretraining data sets a new state-of-the-art in Arabic, Italian, Indonesian, Swahili, and Welsh understanding and commonsense reasoning tasks.
To promote reproducibility, we release our corpus, models, and training pipeline under Open Source Initiative-approved licenses.

\end{abstract}

\section{Introduction}
Multilingual language models have shown remarkable potential for natural language processing~\citep{dubey2024llama, qwen2025qwen25technicalreport, team2024gemma}, yet their development faces a fundamental challenge: the scarcity of high-quality training data across diverse languages~\cite{joshi2020state,kreutzer2022quality}.
While English benefits from extensive, diverse, and well-curated datasets, other languages---even widely spoken ones---struggle to match this standard. 
Current practices of collecting and filtering web data from the Internet lag behind English performance due to the Internet's inherent English-centric nature~\citep{bender2021dangers,imanigooghari-etal-2023-glot500}.

One direction to address the data quality issue is using pretrained language models to generate high-quality synthetic data~\cite{maini2024rephrasing,abdin2024phi}.
However, language model-based data generation is not suitable for multilingual research due to limited language coverage.
For example, one of the current strongest models with multilingual support, \textit{Llama 3.2}~\citep{dubey2024llama}, officially supports fewer than 20 languages.
For low-resource languages like Welsh, the near-absence of well-supported large language models makes data generation practically impossible.
Furthermore, even with proprietary language models such as GPT-4, which might have better coverage, generation at the trillion-token scale is non-trivial.\footnote{The cost would approximately exceed millions of dollars.}

To address the limited language coverage issue, machine translation~(MT) offers a potential solution for creating multilingual datasets using task-specific neural machine translation (NMT) models, with the goal of preserving contextual, idiomatic, and pragmatic nuances across languages.
However, this line of research~\citep{urbizu2023mtrescue,doshi2024translationese} remains underexplored for pretraining.

In this work, we introduce \thedata{}, a large-scale multilingual corpus created by translating a subset of~\textit{FineWeb-edu}~\citep{lozhkov2024fineweb-edu}, a high-quality English dataset, into nine diverse languages using \textit{NLLB-200-1.3B}~\citep{costa2022no}. 
The resulting corpus spans ten languages—Arabic, French, German, Indonesian, Italian, Russian, Spanish, Swahili, Welsh, and English—exceeding $100$B tokens per language, with a total of 1.7 trillion tokens.
%
% In this work, we explore using an MT model
% % rather than prompting large language models, 
% to generate multilingual datasets by translating a trillion-token scale corpus covering ten languages.
% %
% We present a large-scale multilingual corpus~(\thedata{}) created using the~\textit{NLLB-200-1.3B}~\citep{costa2022no} translation model.
% %
% We translate FineWeb-edu, a high-quality English dataset containing 100B tokens, into nine languages representing diverse language families: Arabic, French, German, Indonesian, Italian, Russian, Spanish, Swahili, and Welsh.
% %
% The resulting multilingual corpus contains $100$B+ tokens for each of the ten languages~(including English), totaling $1.7$ trillion tokens.
% %
%
We evaluate the efficiency of \thedata{} by pretraining a $1.3$B-parameter language model.
Contrary to common concerns about text quality when using ``small'' sentence-level NMT models, we demonstrate that it can yield substantial improvements on pretraining performance. 
For instance, \thedata{} improves Swahili by 10\% over the SOTA \textit{Gemma}~\citep{team2024gemma} and enhances Arabic and Italian, surpassing it by 5\% and 2.6\%, respectively.

% ---including issues with text quality and the fact that most models trained at the sentence level.
%
In summary, our contributions are as follows: 
\begin{enumerate}
    \item We translate a high-quality, pretraining-scale English corpus into nine diverse languages, including three medium- and low-resource languages, using a sentence-level NMT model, creating one of the largest machine-generated multilingual datasets to date, \thedata{}, containing 1.7T tokens.
    \item We pretrain \themodel{}, a 1.3B-parameter model, from scratch on our \thedata{} dataset. The model establishes SOTA-level multilingual performance on a wide range of reasoning tasks across nine languages despite using significantly fewer tokens than leading models trained on closed-source data, including \textit{Gemma}, \textit{Qwen2.5}, and \textit{Llama3.2}.

    % Despite using significantly fewer tokens—only 18\% of \textit{Llama3.2}'s training data matches or surpasses SOTA multilingual models trained on closed-source data, including \textit{Gemma}, \textit{Qwen2.5}, and \textit{Llama3.2}, on a wide range of reasoning tasks across nine languages.

    \item We release our corpus, models, and training pipeline under open licenses to advance reproducibility in multilingual NLP.
\end{enumerate}

\section{Pretraining with Machine-translated Multilingual Data}
This section outlines our pipeline for constructing a machine-translated corpus and pretraining a multilingual language model on it.
%
% Our process consists of the following steps. First, we select a high-quality English pretraining dataset. Second, we segment the documents into sentences, translate these sentences into target languages using a sentence-level neural machine translation model, and then reconstruct the documents by concatenating the translated sentences for each target language. Finally, we train a language model from scratch on the resulting multilingual data mixture to validate the effectiveness of the pretraining corpus.
%
Our process consists of the following steps: 
\textbf{(i)} We select a high-quality English pretraining dataset; \textbf{(ii)} We segment English documents into sentences, translate each sentence into target languages using a sentence-level neural machine translation model, and reconstruct the documents by concatenating the translated sentences; \textbf{(iii)} We train a language model from scratch on the resulting multilingual data mixture and validate the effectiveness of the pretraining corpus.
% \begin{enumerate}
%     \item We select a high-quality English pretraining dataset.
    
%     \item We segment English documents into sentences, translate each sentence into target languages using a sentence-level neural machine translation model, and reconstruct the documents by concatenating the translated sentences.

%     \item We train a language model from scratch on the resulting multilingual data mixture to validate the effectiveness of the pretraining corpus.
% \end{enumerate}

\subsection{Pretraining Data Curation}
LLMs are predominantly trained on document-level text data, as demonstrated by prominent model families such as Llama~\citep{dubey2024llama} and Gemma~\citep{team2024gemma}. In keeping with this established approach, we structure our translation pretraining data at the document level. Translation data typically comprises the following key components: source data, target languages, translation model, and a translation approach for composing document-level translations.

\paragraph{Source data}
The quality of a pretraining dataset significantly influences the performance of LLMs. Among high-resource languages, English stands out for its vast linguistic diversity and extensive knowledge base~\citep{joshi2020state, kreutzer2022quality}. This makes it an excellent choice for sourcing high-quality web data for model training.

The \textit{FineWeb-Edu} dataset~\citep{lozhkov2024fineweb-edu}, a subset of \textit{FineWeb}~\citep{penedo2024finewebdatasetsdecantingweb}, has demonstrated its quality and efficacy in the development of LLMs. This dataset, constructed using scalable automated high-quality annotations for educational value, has been instrumental in training both English-centric models like \textit{GPT-2}~\citep{Karpathy2022,Karpathy2024} and multilingual models such as \textit{EuroLLM}~\citep{martins2024eurollm}, and thus emerges as a suitable candidate for our source dataset. It originally consists $1.3$ trillions tokens of educational contents, and our focus is on the sample-$100$BT subset\footnote{\href{https://huggingface.co/datasets/HuggingFaceFW/fineweb-edu}{\texttt{hf.co/datasets/HuggingFaceFW/fineweb-edu}}}, which contains about $100$ billion \textit{GPT-2} tokens randomly sampled from the whole dataset. 

\paragraph{Target Languages}
We focus on $9$ target languages from diverse linguistic families, ensuring broad representation. From the \textit{\textbf{Indo-European}} family, we include Germanic languages: English (en) and German (de); Romance languages: French (fr), Spanish (es), and Italian (it); a Celtic language: Welsh (cy); and a Slavic language: Russian (ru). Additionally, we include languages from distinct families: \textit{\textbf{Afroasiatic}} (Arabic (ar)), \textit{\textbf{Niger-Congo}} (Swahili (sw)), and \textit{\textbf{Austronesian}} (Indonesian (id)). According to~\citet{joshi2020state} and~\citet{ezeani-etal-2019-leveraging}, Indonesian is categorized as a medium-resource language, while Swahili and Welsh are classified as low-resource languages. The remaining languages are considered high-resource. 
% This linguistic diversity enables us to evaluate our approach across a wide range of grammatical structures and typological features.
% To investigate whether the strengths of high-quality English data can be effectively transferred to these nine target languages,  
We translate the $100$BT \textit{FineWeb-Edu} corpus from English into the target languages to bridge resource gaps and transfer the rich knowledge representations encoded in English data, ultimately enhancing the inclusion and representation of these target languages in multilingual LLMs.

% To achieve this, we translate the $100$BT \textit{FineWeb-Edu} corpus from English into these nine target languages. Our goal is to investigate whether the strengths of high-quality English data can be effectively transferred to other languages, including medium- and low-resource ones, through translation. By leveraging translation, we aim to bridge resource gaps and convey the rich linguistic and knowledge representations encoded in English data, ultimately enhancing the inclusion of these languages in multilingual LLMs.

\paragraph{Translation Model}
% @JY: introduce NLLB model, and their variants, why we choose its smallest version 1.3B model, and its MT performance/ranking on ten languages.
Although various models support translation tasks, including neural machine translation (NMT) models and LLMs~\citep{stahlberg2020neural,alves2024tower,martins2024eurollm}. However, exploration of LLMs' translation performance for medium- and low-resource languages remains limited. For example, TowerLLM~\citep{alves2024tower}, an open multilingual LLM designed for translation-related tasks, supports only 10 languages. In contrast, NMT models are generally more accessible and widely adopted for translation, particularly for low-resource languages, due to more than a decade of dedicated development in this field~\citep{stahlberg2020neural}. A prominent example is NLLB-200~\citep{costa2022no}, an open-source suit of models capable of providing high-quality translations for 200 languages. These models are purpose-built to support linguistic diversity and are optimized for robust performance, even for resource-scarce languages.

In our study, we assess whether constructing documents using sentence-level translations can still yield robust pretraining performance for LLMs. We hypothesize that essential linguistic patterns and semantic structures in high-quality source data can be preserved despite translation imperfections---an attribute that could be particularly beneficial for cold-start pretraining in low-resource languages. If successful, this method could greatly expand access to multilingual pretraining data.

Specifically, we begin by segmenting English documents into sentences with the NLTK sentence tokenizer~\citep{bird2009natural} and then translate them into target languages using the NLLB-200 model~\citep{costa2022no}. The translated sentences are reassembled into documents, with their original order and structural elements intact (e.g., newline characters)\footnote{The translation pipeline is detailed in Appendix~\ref{sec:trans_pip}.}. We employ \textit{NLLB-200-1.3B}, the smallest non-distilled model in the NLLB-200 suite, trying to explore the lower bound of translation quality. While this model may not represent the absolute lower bound, it provides a practical reference point for assessing the impact of noisier translations on LLM pretraining.
% Although it may not represent the absolute lower bound of translation quality, it provides a practical reference point for exploring the effects of noisier translations on LLM pretraining. 

% To assess the impact of translation quality on downstream performance, we employ \textit{NLLB-200-1.3B}---the smallest non-distilled model in the NLLB-200 suite. 
% % This choice allows us to try to explore the lower bound of translation quality by incorporating noisier translations, thereby evaluating their suitability for LLM pretraining. 
% Although it may not represent the absolute lower bound of translation quality, it provides a practical reference point for exploring the effects of noisier translations on LLM pretraining. 
% % Moreover, this approach underscores the robustness of LLM pretraining methodologies, highlighting their ability to generalize effectively even when trained on noisier multilingual corpora.

\paragraph{\thedata}
Through the aforementioned approach, we obtain a machine-translated pretraining corpus covering $10$ languages in total. We refer to this multiway parallel translated dataset as \textbf{\thedata{}}. The NLLB translation is performed with a batch size of 4096 and a beam size of 1\footnote{For each target language, the entire translation process is completed within 168 GPU hours using a single GH200 GPU.}. Table~\ref{tab:nllb_statistics} in Appendix~\ref{sec:appendix-nllb-statistics} presents the statistics of the original English data alongside those of the translated components in \thedata{}. To the best of our knowledge, \thedata{} is the largest multiway parallel, document-level multilingual pretraining dataset currently available.

% We translate the English documents from the $100$BT subset of \textit{FineWeb-Edu}\citep{lozhkov2024fineweb-edu} into nine target languages using NLLB-200-1.3B, resulting in a dataset covering ten languages in total. We refer to this multiway parallel translated dataset as \thedata. Sentence-level translation is performed with a batch size of 4096 and a beam size of 1\footnote{For each target language, the entire translation process is completed within 7×24 GPU hours using a single NVIDIA GH200 GPU.}. Table~\ref{tab:nllb_statistics} in Appendix~\ref{sec:appendix-nllb-statistucs} presents the statistics of the original English data alongside those of the translated components in \thedata. To the best of our knowledge, \thedata{} is the largest multiway parallel, document-level multilingual pretraining dataset currently available.

\subsection{Multilingual LM Pretraining}
In this section, we will describe the technical details of pretraining multilingual language models using \thedata{}.

\paragraph{Model Architecture and Hyper-parameters} 
We pretrain a multilingual language model from scratch using \thedata{}, naming it \themodel{}. The architecture and hyperparameter selection of \themodel{} (detailed in Appendix Section~\ref{sec:appendix-model-arch-params}) are informed by insights from the \textit{Llama} family of models and other open-source efforts~\cite{Karpathy2024} to reproduce \textit{GPT-2}. \themodel{} consists of $1.3$B parameters in total. Similar to \textit{GPT-2} reproduction efforts, we adopt a constant learning rate of $6\times 10^{-4}$, a sequence length of up to $2048$ tokens, and a batch size of $2048$ per iteration, resulting in approximately $4$ million tokens processed per iteration.

% We pretrain a mulitlingual language model using \thedata{} from scratch, and name the model with \themodel{}. \themodel{}'s architecture and hyperparameter selection (detailed in Appendix Section~\ref{sec:appendix-model-arch-params}) are based on lessons from the \textit{Llama} family of models and other open-source efforts~\cite{Karpathy2024} to reproduce \textit{GPT-2}. 
% %
% The model has a total of $1.3$B parameters.
% % , which is comparable to a wide range of small language models with sizes ranging from $1$B to $2$B parameters.
% Similar to \textit{GPT-2} reproduction efforts, we use a constant learning rate of $6\times 10^{-4}$ and a sequence length of up to $2048$ tokens, with a batch size of $2024$ per iteration, resulting in approximately $4$ million tokens processed per iteration. 

\paragraph{Tokenization} 
% In this work, we focus on a total of ten languages: Arabic (ar), English (en), French (fr), German (de), Italian (it), Russian (ru), Spanish (es), Indonesian (id), Swahili (sw), and Welsh (cy). 
\citet{alves2024tower} extends the multilingual capabilities of \textit{Llama2} models~\citep{touvron2023llama2}, demonstrating that the \textit{Llama2} tokenizer remains a practical choice for ensuring efficiency across diverse languages. Building on their findings, we use the \textit{Llama2} tokenizer in our experiments. For non-Latin languages, such as Arabic and Russian, it tokenizes the text while representing it using Unicode-based embeddings.

\paragraph{Pretraining Data} 
% Although \thedata{} is one of the largest multiway parallel, document-level multilingual datasets available, it is not designed to provide aligned translation pairs for pretraining. 
During pretraining, we use the same pretraining setup as GPT-family models, where documents are randomly sampled from the corpus.
As a result, the likelihood of the same document appearing in different languages within a single batch is very low. For clarity, Table~\ref{tab:training-example} presents an example of our pretraining data.
% Although \thedata{} is one of the largest multiway parallel, document-level multilingual datasets available, we do not intend to create aligned translation pairs for pretraining. Instead, we adopt a monolingual pretraining setup, where documents are randomly fetched from the corpus. This approach means that the probability of the same document appearing in different languages within a single batch is very low. For clarity, we illustrate an example of our pretraining data in Table~\ref{tab:training-example}.
\begin{table}[!t]
\small
\centering
\resizebox{1.0\columnwidth}{!}{%
\begin{tabular}{l}
\toprule
\textbf{GPT2-style Training Sequence} \\
\midrule
\texttt{<random-fr-doc><eos><random-en-doc><eos>....<random-ru-doc>} \\
\bottomrule
\end{tabular}
}
\caption{An illustration of our pretraining sequence containing multiple non-parallel documents.}
\label{tab:training-example}
\end{table}

\paragraph{Framework and Training} \themodel{} is trained using the \textit{Megatron-LM} framework~\citep{shoeybi2019megatron} from scratch with an accelerated attention implementation~\citep{dao2023flashattention}.  
The training is done on the
%UK Isambard-AI 
NVIDIA GH200 cluster~\citep{mcintosh2024isambard} for $8,366$ GPU hours.
% The training is done on a NVIDIA GH200 cluster for $3,187$ GPU hours.
Our pretraining dataset \thedata{} is created on a balanced basis, so we do not perform any up-sampling for specific languages. Our pretraining over \thedata{} processes approximately $1.5$T tokens, which is nearly one epoch of the pretraining data. Similar to the observation of \citet{muennighoff2024datascaling}, we observe no degradation in performance over the validation set during this process.

\begin{table*}[!t]
\centering
\scriptsize
\setlength{\tabcolsep}{4pt} 
\resizebox{0.95\textwidth}{!}{%
\begin{tabular}{l|ccccccc|c|cc|ccc}
\toprule
& \multicolumn{7}{c|}{\textit{High}} & \textit{Medium} & \multicolumn{2}{c|}{\textit{Low}} & \multicolumn{3}{c}{\textbf{Average}} \\
 & ar & en & fr & de & it & ru & es & id & sw & cy & All & Non-eng & High \\
\hline
\multicolumn{14}{@{}l@{}}{\textit{\textbf{English LLMs}}} \\
Pythia (1.4B) & 33.21 &  54.63 & 40.14 & 36.23 & 35.14 & 38.32 & 39.91 & 36.83 & 36.90 & 31.57  & 38.29 & 36.47 & 39.65 \\
TinyLlama (1.1B) & 32.86 & 57.18 & 44.13 & 37.06 & 36.79 & 40.97 & 42.38 & 36.13 & 36.69 & 31.53  & 39.57 & 37.62 & 41.62 \\
\hline
\multicolumn{14}{@{}l@{}}{\textit{\textbf{Multilingual LLMs}}} \\
mGPT (1.3B) & 32.90 & 45.38 & 40.27 & 34.54 & 34.89 & 40.53 & 38.88 & 39.47 & 38.98 & 31.14 & 37.70 & 36.84 & 38.20 \\
BLOOM (1.1B) & 34.97 & 50.94 & 41.80 & 34.62 & 33.69 & 37.56 & 42.69 & 43.23 & 37.09 & 31.57 & 38.82 & 37.47 & 39.47 \\
Llama3.2 (1.3B) & 34.78 & 58.16 & 44.10 & 39.73 & 40.93 & 45.38 & 44.15 & 44.67 & 38.30 & 31.84 & 42.20 & 40.43 & 43.89 \\
Qwen2 (1.5B) & 35.61 & \underline{61.07} & 47.40 & 40.79 & 42.61 & \underline{47.95} & \underline{47.40} & 45.93 & 38.89 & 32.23 & 43.99 & 42.09 & 46.12 \\
Qwen2.5 (1.5B) & 37.35 & \underline{\cellcolor{yellow!30}62.93} & \underline{48.69} & 40.49 & 43.10 & \underline{47.01} & \underline{48.41} & 46.17 & 37.98 & 31.78 & \underline{44.39} & \underline{42.33} & \underline{46.85} \\
Gemma (2.6B) & 37.26 & \underline{62.39} & \underline{\cellcolor{yellow!30}49.78} & \underline{\cellcolor{yellow!30}44.27} & \underline{44.57} & \underline{\cellcolor{yellow!30}48.60} & \underline{\cellcolor{yellow!30}49.35} & \underline{48.27} & \underline{40.18} & \underline{32.28} & \underline{\cellcolor{yellow!30}45.70} & \underline{\cellcolor{yellow!30}43.84} & \underline{48.03} \\
\hline
\multicolumn{14}{@{}l@{}}{\textit{\textbf{Language-Specific LLMs}}} \\
AfriTeVa (1B) & & 37.70 & & & & & & & \underline{40.60} & & &\\
BritLLM (3B) & & 60.45 & & & & & & & & \underline{37.07} & & &\\
CroissantLLM (1.3B) & & 53.31 & 45.73 & & & & & & & & &\\
EuroLLM (1.7B) & \underline{38.88} & 57.98 & \underline{47.85} & \underline{42.07} & \underline{\cellcolor{yellow!30}47.56} & 46.71 & 47.07 & &  &  &  & & \underline{46.87}\\
% GPT-fr (1B) & & 0.3887 & 0.3736 & & & & & & & & & &\\
Jais-family-1p3b (1.3B) & \underline{\cellcolor{yellow!30}39.97} & 56.28 & & & & & & & & & & &\\
Sailor (1.8B) & & 55.40 & & & & & & \underline{48.45} & & & & & \\
Sailor2 (1B) & & 54.61 & & & & & & \underline{\cellcolor{yellow!30}49.44} & & & & & \\
\hline
\multicolumn{14}{@{}l@{}}{\textit{\textbf{Ours}}} \\
TransWebLLM (1.3B) & \underline{39.30} & 56.30 & 46.11 & \underline{41.01} & \underline{45.75} & 46.38 & 45.27 & 47.54 & \underline{\cellcolor{yellow!30}44.44} & \underline{\cellcolor{yellow!30}38.69} & \underline{45.08} & \underline{43.83} & 45.73 \\
\bottomrule
\end{tabular}}
% \caption{LLM performance across ten languages, categorized by resource availability and measured in accuracy, and they are the averaged scores across benchmarks detailed in Section~\ref{sec:benchmarks} per language. The last three columns report average results for all languages (All), non-English languages (Non-Eng), and high-resource languages (High). The top three models are underlined and best scores are highlighted.}
\caption{LLM performance across ten languages, grouped by resource availability and measured in accuracy. Scores represent the average accuracy across benchmarks detailed in Section~\ref{sec:benchmarks}. The last three columns report mean results for all languages (All), non-English languages (Non-Eng), and high-resource languages (High). The top three models are underlined, and the best score for each language is highlighted.}
\label{tab:main_result}
\end{table*}

\section{Experiments}
This section presents our evaluation of model performance across various multilingual benchmarks.

\subsection{Evaluation Benchmark Datasets}
\label{sec:benchmarks}
Our evaluation includes English benchmarks and assessments for the nine non-English languages, focusing on natural language understanding and common-sense reasoning. All benchmarks used are publicly available and open-source, ensuring the transparency and reproducibility of our results\footnote{All evaluations are conducted using~\url{https://github.com/EleutherAI/lm-evaluation-harness}.}. 

Specifically, our evaluation framework includes the following tasks: \textbf{ARC}~\citep{clark2018think,lai2023okapi, bayes2024uhura}: grade-school level multiple-choice science questions; \textbf{Hellaswag}~\citep{zellers2019hellaswag,lai2023okapi}: common-sense reasoning benchmarks for contextually appropriate sentence endings prediction; \textbf{PAWS-X}~\citep{yang-etal-2019-paws}: a cross-lingual adversarial dataset for paraphrase identification, sourced from English Wikipedia and Quora; \textbf{PIQA}~\citep{Bisk2020}: physical commonsense reasoning benchmarks; \textbf{SciQ}~\citep{SciQ}: an multiple-choice question-answering dataset in scientific topics; \textbf{TruthfulQA}~\citep{lin2021truthfulqa, bayes2024uhura}: question-answering evaluation tasks for truthfulness and factual accuracy of model responses;
\textbf{XCOPA}~\citep{ponti2020xcopa}: across-lingual adaptation of COPA~\citep{roemmele2011choice} for transfer commonsense reasoning evaluation; \textbf{XNLI}~\citep{conneau2018xnli}: an multilingual extension of~\citet{williams-etal-2018-broad}, assessing textual entailment prediction; \textbf{XStoryCloze}~\citep{lin2021few}: an multilingual adaptation of~\citet{mostafazadeh2016corpus} for assessing cross-lingual narrative understanding by predicting story endings; \textbf{XWinograd}~\citep{tikhonov2021s}: a cross-lingual adaptation of the Winograd Schema challenge\footnote{\url{https://cs.nyu.edu/~davise/papers/WinogradSchemas/WS.html}} for coreference resolution evaluation.
% \textbf{ARC} \citep{clark2018think,lai2023okapi, britllm2024}: Grade-school level multiple-choice science questions.
% \textbf{Hellaswag}~\citep{zellers2019hellaswag,lai2023okapi, britllm2024}: Common-sense reasoning benchmarks for  contextually appropriate sentence endings prediction.
% \textbf{PAWS-X}~\citep{yang-etal-2019-paws}: A cross-lingual adversarial dataset for paraphrase identification, sourced from English Wikipedia and Quora.
% \textbf{PIQA}~\citep{Bisk2020, britllm2024}: Physical commonsense reasoning benchmarks.
% \textbf{SciQ}~\citep{SciQ}: A multiple-choice question-answering dataset in scientific topics.
% \textbf{TruthfulQA}~\citep{lin2021truthfulqa, bayes2024uhura, britllm2024}: Question-Answering evaluation tasks for truthfulness and factual accuracy of model responses.
% \textbf{XCOPA}~\citep{ponti2020xcopa}: A cross-lingual adaptation of COPA~\citep{roemmele2011choice} for transfer commonsense reasoning evaluation.
% \textbf{XNLI}~\citep{conneau2018xnli,britllm2024}: An multilingual extension of~\citet{williams-etal-2018-broad}, assessing textual entailment prediction. 
% \textbf{XStoryCloze}~\citep{lin2021few}: An multilingual adaptation of~\citet{mostafazadeh2016corpus} for assessing cross-lingual narrative understanding by predicting story endings
% \textbf{XWinograd}~\citep{tikhonov2021s}: A cross-lingual adaptation of the Winograd Schema challenge\footnote{\url{https://cs.nyu.edu/~davise/papers/WinogradSchemas/WS.html}} for coreference resolution evaluation.
However, not all ten languages have all the above benchmarks publicly available. The specific evaluation datasets used for each language are detailed in Table~\ref{tab:specific_benchmarks} in Appendix~\ref{sec:appendix-benchmarks}\footnote{For Welsh evaluation, we use the \textit{BritEval} benchmarks, \url{https://llm.org.uk/}.}. All benchmarks are evaluated using a standard $5$-shot setting, with results reported in terms of accuracy.

\subsection{Baselines}
We benchmark our \themodel{}, trained with \thedata{}, in a $5$-shot setting, against diverse open-source multilingual and monolingual LLMs with comparable parameter sizes, different multilingual pretraining mixtures and data sources.

Our \textit{\textbf{multilingual LLM baselines}} include:
mGPT ($1.3$B)~\citep{shliazhko2022mgpt},
BLOOM ($1.1$B)~\citep{le2023bloom},
Llama3.2 ($1.3$B)~\citep{dubey2024llama},
Qwen2 ($1.5$B)~\citep{yang2024qwen2technicalreport}, Qwen2.5 ($1.5$B)~\citep{qwen2025qwen25technicalreport}, 
and Gemma ($2.6$B)~\citep{team2024gemma}.
Additionally, we compare against \textit{\textbf{language-specific LLM baselines}}:
Afriteva\_v2\_large ($1$B)~\citep{oladipo2023better} for Swahili,
BritLLM ($3$B)\footnote{
\href{https://hf.co/britllm/britllm-3b-v0.1}{\texttt{hf.co/britllm/britllm-3b-v0.1}}
% \url{https://huggingface.co/britllm/britllm-3b-v0.1}
} for Welsh,
CroissantLLM ($1.3$B)~\citep{faysse2024croissantllm} for French,
EuroLLM ($1.7$B)~\citep{martins2024eurollm} for Arabic, French, German, Italian, Russian, and Spanish, 
Jais-family-1p3b ($1.3$B)~\citep{sengupta2023jais} for Arabic,
Sailor ($1.8$B)~\citep{dou2024sailor} and Sailor2 ($1$B)~\citep{sailor2report} for Indonesian. 
Furthermore, we include two \textit{\textbf{English-centric baselines}} in our evaluation:
TinyLlama ($1.1$B)~\citep{zhang2024tinyllama},
Pythia ($1.4$B)~\citep{biderman2023pythia}. An overview of baseline models and our \themodel{} has been shown in Table~\ref{tab:model_comparison} in Appendix~\ref{sec:appendix-baseline}.

\subsection{Main results of \themodel{}}
Table~\ref{tab:main_result} presents the average performance of \themodel{} across benchmark datasets for each language, comparing it to baseline models across all 10 languages. The last three columns of Table~\ref{tab:main_result} summarize the overall average performance for (i) All languages, (ii) Non-English languages, and (iii) High-resource languages. Generally, \themodel{} consistently ranks among the top three models in terms of average performance across \textit{\textbf{all languages}} and \textit{\textbf{non-English languages}}, with the accuracy scores of $45.08$ and $43.83$, respectively. It significantly outperforms multilingual LLMs such as \textit{mGPT}, \textit{BLOOM}, \textit{Llama3.2}, \textit{Qwen2}, and \textit{Qwen2.5} of similar model size, and achieves comparable results with ~\textit{Gemma}, despite \textit{Gemma} having twice the model size of \themodel{}. 

For \textit{\textbf{high-resource languages}}, \themodel{} ranks among the top three models for Arabic, German, and Italian. Additionally, it outperforms \textit{Llama3.2} on average ($45.73$ vs. $43.89$), despite the latter being trained on 9T tokens—an order of magnitude more than the data used for training \themodel{}. A similar trend is observed in French evaluation when comparing \themodel{} to \textit{CroissantLLM}. \themodel{} outperforms \textit{CroissantLLM} ($46.11$ vs. $45.73$), even though the latter is trained on 3T tokens, with half of them in French. In contrast, \themodel{} is trained on $1.5$T tokens, with less than $10$\% ($150$B) in French, primarily consisting of translated data.

More remarkable results are observed in \textit{\textbf{medium- and low-resource languages}}. For Indonesian, \themodel{} outperforms both \textit{Qwen2} and \textit{Qwen2.5}, despite the latter two being trained on $7$T and $18$T tokens. The most notable gains are seen in Swahili and Welsh, where \themodel{} ranks first among all baselines, achieving accuracy scores of $44.44$ and $38.69$, outperforming \textit{Gemma} ($2.6$B) and \textit{BritLLM} ($3$B).
These results suggest that training with translation data can be a viable cold-start strategy for pretraining LLMs in medium- and low-resource languages.

For \textit{\textbf{English}}, \themodel{} performs slightly worse than \textit{TinyLlama} but outperforms \textit{CroissantLLM}. Despite being trained on significantly less English data—only $150$B tokens compared to \textit{TinyLlama}’s $3$T and \textit{CroissantLLM}’s $1.5$T—\themodel{} achieves competitive performance, underscoring the importance of the high-quality English source. 

\section{Discussion and Ablations}
In this section, we explore (1) the impact of LLM-generated translation data on pretraining performance and (2) the effects of additional data sources, including general web data and specialized datasets such as rephrased synthetic text, code, and instruction data. We address these questions through a series of ablation experiments.

\subsection{Pretraining with Translation Data generated from an LLM}
Recent studies~\citep{alves2024tower, martins2024eurollm} show that LLMs can effectively support translation tasks, raising the question: \textit{How does pretraining performance differ when using a translation corpus generated by an LLM versus an NMT model like NLLB, used in \themodel{}}?

\citet{dubey2024llama} highlighted Mistral’s potential for multilingual NLP, while \citet{moslem2023fine} and \citet{kocmi2024preliminary} demonstrated its effectiveness in machine translation. Based on these insights, we use \textit{Mistral-7B-Instruct-v0.1}\footnote{\href{https://hf.co/mistralai/Mistral-7B-Instruct-v0.1}{\texttt{hf.co/mistralai/Mistral-7B-Instruct-v0.1}}} for translation, and we focus on English, French, German, and Spanish for this ablation. Details on data generation are provided in the Appendix~\ref{sec:appendix-cuatrollm}. A key difference between Mistral- and NLLB-generated translations lies in their approach to text segmentation. Mistral is prompted to translate chunked documents, better preserving contextual coherence, whereas NLLB translates at the sentence level, which may result in inconsistencies in document-level fluency and cohesion. 

Due to computational capacity, we translate $64$B tokens from the sample-$100$BT subset of \textit{FineWeb-Edu}. We then pretrain an LLM from scratch using the same training framework as \themodel{}, naming it \cuatrollm{}. For a fair comparison, we extract the corresponding NLLB-translated data for French, German, and Spanish from \thedata{} and retrain \themodel{} using this subset, referring to the resulting model as \themodel{}-4. 

Both \cuatrollm{} and \themodel{}-4 are evaluated on benchmark datasets for English, French, German, and Spanish, as introduced in Section~\ref{sec:benchmarks}, with results presented in Table~\ref{tab:main_mistral_nllb_compare} in Appendix~\ref{sec:appendix-cuatrollm}. Across four languages, \themodel{} and \cuatrollm{} perform similarly ($46.65$ vs. $47.16$), both surpassing \textit{mGPT}, \textit{BLOOM}, and \textit{Llama3.2} on average. This suggests that the choice of translation method has a limited impact on these specific downstream evaluations. However, a key advantage of the NLLB-200 model---with support for 200 languages---is its scalability and efficiency, making it a more viable choice for expanding multilingual pretraining to a broader range of languages.

\subsection{Beyond Pretraining with Translation Data}
\label{sec:beyond}
In this section, we assess whether adding specialized data provide benefits beyond pretraining with machine-translated data.

\begin{table*}[!t]
\centering
\scriptsize
\setlength{\tabcolsep}{4pt} 
\resizebox{0.95\textwidth}{!}{%
\begin{tabular}{l|ccccccc|c|cc|ccc}
\toprule
& \multicolumn{7}{c|}{\textit{High}} & \textit{Medium} & \multicolumn{2}{c|}{\textit{Low}} & \multicolumn{3}{c}{\textbf{Average}} \\
 & ar & en & fr & de & it & ru & es & id & sw & cy & All & Non-eng & High \\
\hline
\multicolumn{14}{@{}l@{}}{\textit{\textbf{Multilingual LLMs}}} \\
Llama3.2 (1.3B) & 34.78 & 58.16 & 44.10 & 39.73 & 40.93 & 45.38 & 44.15 & 44.67 & 38.30 & 31.84 & 42.20 & 40.43 & 43.89 \\
Qwen2 (1.5B) & 35.61 & 61.07 & 47.40 & 40.79 & 42.61 & 47.95 & 47.40 & 45.93 & 38.89 & 32.23 & 43.99 & 42.09 & 46.12 \\
Qwen2.5 (1.5B) & 37.35 & \cellcolor{yellow!30}62.93 & 48.69 & 40.49 & 43.10 & 47.01 & 48.41 & 46.17 & 37.98 & 31.78 & 44.39 & 42.33 & 46.85 \\
Gemma (2.6B) & 37.26 & 62.39 & \cellcolor{yellow!30}49.78 & \cellcolor{yellow!30}44.27 & 44.57 & \cellcolor{yellow!30}48.60 & \cellcolor{yellow!30}49.35 & 48.27 & 40.18 & 32.28 & 45.70 & 43.84 & \cellcolor{yellow!30}48.03 \\
\hline
\multicolumn{14}{@{}l@{}}{\textit{\textbf{Language-Specific LLMs}}} \\
AfriTeVa (1B) & & 37.70 & & & & & & & 40.60 & & &\\
BritLLM (3B) & & 60.45 & & & & & & & & 37.07 & & &\\
CroissantLLM (1.3B) & & 53.31 & 45.73 & & & & & & & & &\\
EuroLLM (1.7B) & 38.88 & 57.98 & 47.85 & 42.07 & 47.56 & 46.71 & 47.07 & &  &  &  & & 46.87\\
Jais-family-1p3b (1.3B) & \cellcolor{yellow!30}39.97 & 56.28 & & & & & & & & & & &\\
Sailor (1.8B) & & 55.40 & & & & & & 48.45 & & & & & \\
Sailor2 (1B) & & 54.61 & & & & & & 49.44 & & & & & \\
\hline
\multicolumn{14}{@{}l@{}}{\textit{\textbf{Ours}}} \\
TransWebLLM (1.3B) & 39.30 & 56.30 & 46.11 & 41.01 & 45.75 & 46.38 & 45.27 & 47.54 & \cellcolor{yellow!30}44.44 & 38.69 & 45.08 & 43.83 & 45.73 \\
TransWebLLM-web (1.3B) & 39.82 & 56.18 & 46.83 & 41.92 & \cellcolor{yellow!30}47.01 & 46.22 & 46.92 & \cellcolor{yellow!30}49.75 & 44.40 & \cellcolor{yellow!30}40.09 & \cellcolor{yellow!30}45.91 & \cellcolor{yellow!30}44.77 & 46.41 \\
$\Delta$ Gain & +0.52 & -0.12 & +0.72 & +0.91 & +1.26 & -0.16 & +1.65 & +2.21 & -0.04 & +1.40 & +0.83 & +0.94 & +0.68 \\
\bottomrule
\end{tabular}}
\caption{Performance comparison between \themodel{} and \themodelweb{} across ten languages. The last row ($\Delta$ Gain) shows the performance difference, with positive values indicating improvements of \themodelweb{} over \themodel{}. The best score for each language is highlighted.}
\label{tab:main_real}
\end{table*}

% \begin{table}[!t]
% \centering
% \resizebox{\columnwidth}{!}{%
% \begin{tabular}{l|ccc}
% \toprule
% \textbf{Model} & \textbf{\# tokens} & \textbf{Method} & \textbf{Data} \\
% \midrule
% \themodel & 1.5T   &  Train from scratch & \thedata\\
% \midrule
% \themodelweb{} & +90B &  {\makecell{Continue train \\on \themodel}} & {\makecell{\thedata{} \\+ Real web data}} \\
% \midrule
% \themodelmc{} & +40B & {\makecell{Continue train \\ on \themodelweb{}}} & {\makecell{\thedata{} \\+ Real web data \\+ MC synthetic data}} \\
% \midrule
% \themodelcool{} & +20B & {\makecell{Continue train \\ on \themodelmc{}}} & {\makecell{\thedata{} \\+ Real web data \\+ MC synthetic data \\ + Cooldown Data}}\\
% \bottomrule
% \end{tabular}
% }
% \caption{Models used for different data impact experiments.}
% \label{tab:data-impact-model-details}
% \end{table}

\begin{table}[!t]
\centering
\Large % Increase font size for the entire table
\resizebox{\columnwidth}{!}{%
\begin{tabular}{l|ccc}
\toprule
\textbf{Model} & \textbf{\# tokens} & \textbf{Method} & \textbf{Data} \\
\midrule
\themodel & 1.5T   &  Train from scratch & \thedata\\
\midrule
\themodelweb{} & +90B &  {\makecell{Continue train on \\ \themodel}} & {\makecell{\thedata{} \\+ Real web data}} \\
\midrule
% \themodelmc{} & +40B & {\makecell{Continue train \\ on \themodelweb{}}} & {\makecell{\thedata{} \\+ Real web data \\+ MC synthetic data}} \\
% \midrule
\themodelcool{} & +60B & {\makecell{Continue train on \\ \themodelweb{}}} & {\makecell{\thedata{} \\+ Real web data \\+ MC synthetic data \\ + Cooldown Data}}\\
\bottomrule
\end{tabular}
}
\caption{Models used in data impact ablations.}
\label{tab:data-impact-model-details}
\end{table}

\subsubsection{Impact of General Web Data}
\label{sec:real_web}
Given that \thedata{} is largely based on educational contents, which is a highly specialized domain, we investigate whether multilingual reasoning capabilities can be further enhanced by incorporating general web data.

We construct the general web dataset by sampling English, French, German, Italian, and Spanish data from \textit{RedPajama-v2} (\textit{RPv2})~\citep{NEURIPS2024_d3449733}; Arabic, Russian, and Indonesian from \textit{mC4}~\citep{xue-etal-2021-mt5}; Swahili from \textit{Wura}~\citep{oladipo2023better}; and Welsh from \textit{CC100}~\citep{wenzek-etal-2020-ccnet}. For \textit{RPv2}, we filter each subset using its built-in quality signals\footnote{Details are introduced in Appendix~\ref{sec:appendix-rpv2-sampling}.}; for~\textit{mC4}, we apply random sampling. Given the limited availability of Swahili and Welsh data in \textit{Wura} and \textit{CC100}, we include their entire datasets. We balance the general web data by sampling an equal number of tokens per language, upsampling Indonesion, Swahili and Welsh as needed to match their proportions in \thedata{}. We then merge it with \thedata{} at an nearly $1:1$ ratio for continued pretraining. Building on \themodel{}, we extend training for an additional $20,800$ steps, processing approximately $90$B tokens during this phase, with general web data accounting for only around $45$B tokens (less than 3\% of the total). We refer to this continued pretraining model as \themodelweb{}, as detailed in Table~\ref{tab:data-impact-model-details}.

\paragraph{Understanding and Reasoning Evaluation}
The 5-shot evaluation results of \themodelweb{} on multilingual understanding and commonsense reasoning benchmarks (as detailed in Section~\ref{sec:benchmarks}) are in Table~\ref{tab:main_real}. \themodelweb{} demonstrates a significant improvement over \themodel{}, achieving higher average performance in general. Across all languages and non-English languages, \themodelweb{} ranks as the top-performing model among all LLMs ($45.91$ and $44.77$). For high-resource languages, \themodelweb{} achieves performance comparable to \textit{EuroLLM}, despite the latter being trained on 4T tokens. Notably, for Indonesian, \themodelweb{} emerges as the best-performing model ($49.75$), surpassing Southeast Asian-specific baselines such as \textit{Sailor} and \textit{Sailor2}. These results highlight the benefits of incorporating even a limited amount of general web data during continued pretraining for multilingual understanding and reasoning tasks.

\paragraph{Linguistic Proficiency Evaluation}
Beyond understanding and reasoning tasks, we also evaluate the model’s linguistic proficiency, focusing on its ability to understand and generate coherent, grammatically accurate sentences. \citet{faysse2024croissantllm} introduced the \textit{fr-grammar} and \textit{fr-vocabulary} test sets in French to assess models' grammar and vocabulary capabilities through structured language evaluations. We test both \themodel{} and \themodelweb{} on these benchmarks in a 5-shot setting to measure their proficiency in French linguistic competence. As shown in Table~\ref{tab:french_ling}, \themodelweb{} outperforms \themodel{} by nearly $10$ accuracy points ($74.79$ vs. $65.13$), demonstrating that even a small addition of general web data ($45$B) in continued pretraining can significantly enhance linguistic proficiency.

\begin{table}[!t]
\centering
\small
\resizebox{0.44\textwidth}{!}{%
\begin{tabular}{l|cc|c}
\toprule
\textbf{Model} & \textit{\textbf{fr-grammar}} & \textit{\textbf{fr-vocab}} & \textbf{Avg.} \\
\midrule
\multicolumn{4}{@{}l@{}}{\textit{\textbf{Baselines}}} \\
mGPT & 73.95 & 70.59 & 72.27 \\
BLOOM & 79.83 & 74.79 & 77.31 \\
Llama3.2 & 76.47 & 75.63 & 76.05 \\
EuroLLM$^{*}$ & 79.83 & 78.99 & 79.41 \\
% Qwen2 & 71.43 & 73.95 & 72.69 \\
Qwen2.5 & 71.43 & 73.95 & 72.69 \\
Gemma & 73.11 & 72.27 & 72.69 \\
CroissantLLM$^{*}$ & 79.83 & 78.15 & 78.99 \\
\hline
\multicolumn{4}{@{}l@{}}{\textit{\textbf{Ours}}} \\
TransWebLLM & 67.23 & 63.03 & 65.13 \\
TransWebLLM-web & 73.11 & 76.47 & 74.79 \\
\bottomrule
\end{tabular}}
\caption{French grammar and vocabulary proficiency evaluation, measured in accuracy. Models with $^{*}$ denote regional models trained with support for French.}
\label{tab:french_ling}
\end{table}

\begin{table}[!t]
\centering
\small
\resizebox{0.44\textwidth}{!}{%
\begin{tabular}{l|cc|c}
\toprule
\textbf{Model} & \textbf{\textit{colloquial}} & \textbf{\textit{standard}} & \textbf{Avg.} \\
\midrule
\multicolumn{4}{@{}l@{}}{\textit{\textbf{Baselines}}} \\
mGPT & 54.56 & 53.49 & 54.03 \\
BLOOM & 55.10 & 54.38 & 54.74 \\
Llama3.2 & 52.42 & 52.59 & 52.51 \\
Qwen2.5 & 52.59 & 54.03 & 53.31 \\
Gemma & 54.03 & 55.99 & 55.01 \\
Sailor$^{*}$ & 57.60 & 65.47 & 61.54 \\
Sailor2$^{*}$ & 58.86 & 66.37 & 62.62 \\
\hline
\multicolumn{4}{@{}l@{}}{\textit{\textbf{Ours}}} \\
TransWebLLM & 48.12 & 49.55 & 48.84 \\
TransWebLLM-web & 55.46 & 59.75 & 57.61 \\
\bottomrule
\end{tabular}}
\caption{COPAL-ID evaluation, measured in accuracy. Models with $^{*}$ denote regional models trained with support for Indonesian.}
\label{tab:local_culture_reasoning}
\end{table}

\begin{table*}[!t]
\centering
\scriptsize
\setlength{\tabcolsep}{4pt} 
\resizebox{0.95\textwidth}{!}{%
\begin{tabular}{l|ccccccc|c|c|cc}
\toprule
& \multicolumn{7}{c|}{\textit{High}} & \textit{Medium} & \textit{Low} & \multicolumn{2}{c}{\textbf{Average}} \\
 & ar & en & fr & de & it & ru & es & id & sw & All & High \\
\hline
\multicolumn{11}{@{}l@{}}{\textit{\textbf{Multilingual LLMs}}} \\
mGPT (1.3B) & 25.02 & 25.27 & 26.10 & 24.05 & 25.70 & 25.48 & 25.64 & 25.10 & 24.11 & 25.16 & 25.32 \\
BLOOM (1.1B) & 26.36 & 26.25 & 26.65 & 26.51 & 27.25 & 26.76 & 26.09 & 25.86 & 26.61 & 26.48 & 26.55  \\
Llama3.2 (1.3B) & 27.72 & 31.17 & 27.69 & 27.94 & 27.67 & 27.54 & 28.19 & 27.86 & 26.39 & 28.02 & 28.27 \\
% Qwen2 (1.5B) & 40.36 & 54.92 & 47.43 & 45.01 & 45.73 & 43.87 & 47.64 & 45.38 & 30.35 & 44.52 & 46.42 \\
Qwen2.5 (1.5B) & \underline{42.24} & \underline{59.37} & \underline{50.43} & \underline{48.23} & \underline{49.17} & \underline{46.38} & \underline{51.89} & \underline{47.16} & \underline{30.62} & \underline{47.28} & \underline{49.67} \\
Gemma (2.6B) & \underline{31.96} & \underline{40.97} & \underline{34.52} & \underline{35.34} & \underline{34.45} & \underline{32.83} & \underline{35.38} & 32.58 & \underline{30.90} & \underline{34.33} & \underline{35.06} \\
\hline
\multicolumn{11}{@{}l@{}}{\textit{\textbf{Language-Specific LLMs}}} \\
AfriTeVa (1B) & & 26.87 & & & & & & & 26.93 & &  \\
CroissantLLM (1.3B) & & 25.35 & 25.36 & & & & & & & & \\
EuroLLM (1.7B) & 26.23 & 27.13 & 26.79 & 26.47 & 26.25 & 27.61 & 26.37 & &  &  & 26.69  \\
Jais-family-1p3b (1.3B) & 25.94 & 25.06 & & & & & & & & &  \\
Sailor (1.8B) & & 28.62 & & & & & & 26.39 & & &  \\
Sailor2 (1B) & & \underline{37.03} & & & & & & \underline{33.34} & & &  \\
\hline
\multicolumn{11}{@{}l@{}}{\textit{\textbf{Ours}}} \\
TransWebLLM (1.3B) & 26.63 & 24.66 & 25.69 & 25.46 & 25.32 & 26.42 & 26.21 & 25.28 & 25.35 & 25.67 & 25.77  \\
TransWebLLM-web (1.3B) & 26.49 & 26.41 & 26.84 & 26.11 & 26.16 & 26.56 & 26.58 & 26.68 & 26.48 & 26.48 & 26.45 \\
% TransWebLLM-mcSyn (1.3B) & 30.52 & 34.53 & 32.99 & 32.69 & 33.17 & 32.08 & 33.01 & 32.84 & 30.93 & 32.53 & 32.71 \\
TransWebLLM-cool (1.3B) & \underline{30.44} & 34.26 & \underline{32.58} & \underline{32.27} & \underline{31.95} & \underline{32.50} & \underline{32.53} & \underline{33.18} & \underline{31.11} & \underline{32.31} & \underline{32.36} \\
\bottomrule
\end{tabular}}
\caption{Evaluation on Global-MMLU full sets~\citep{singh2024global}, measured in accuracy. The rightmost columns report the average scores across all languages (All) and high-resource languages (High). Top 3 models are underlined.}
\label{tab:main_mmlu}
\end{table*}

\paragraph{Reasoning Evaluation for Local Culture}
Local culture reasoning provides a natural representation of causal reasoning within specific cultural contexts. COPAL-ID~\citep{wibowo2023copal} is an Indonesian causal commonsense reasoning dataset, written by native speakers from scratch with standard Indonesian and Jakartan Indonesian, a widely spoken dialect in daily conversations. We evaluate both \themodel{} and \themodelweb{} on this benchmark in a 5-shot setting to assess their ability to reason within the Indonesian cultural sphere. As shown in Table~\ref{tab:local_culture_reasoning}, \themodelweb{} improves Indonesian cultural reasoning by over an averaged 8 accuracy points ($57.61$ vs. $48.84$) by incorporating a limited amount of general web data ($45$B) in continued pretraining on \themodel{}. It surpasses all LLM baselines except \textit{Sailor} and \textit{Sailor2}, which have been specifically trained for Indonesian. 

\subsubsection{Impact of Special Data}
\citet{yang2023rethinking} shows that rephrasing MMLU~\citep{hendrycks2021measuring} samples enhances model reasoning performance across various domains. Motivated by these findings, we explore the impact of \textbf{\textit{rephrased synthetic data}} on \themodel{}. Instead of rephrasing MMLU test cases~\citep{yang2023rethinking}, we rephrase English web data into a multiple-choice (MC) style using an LLM, aligning with reasoning structure while maintaining its open-ended nature. 
% We then incorporate this synthetic data into our multilingual pretraining corpus, assessing if rephrased English MC data can enhance multilingual reasoning through cross-lingual transfer. 
We extract $10$BT English data from SlimPajama~\citep{cerebras2023slimpajama}, generate $8$BT MC synthetic data using \textit{Mistral-7B-Instruct-v0.1}\footnote{We use the prompt template as "\textit{Write multiple-choice questions and answers based on the document: [doc]}".}, and upsample and integrate it into \thedata{} with general web data, ensuring MC data constitutes 5\% of the corpus. Given the improved performance of \themodelweb{}, we continue pretraining for 9,000 steps, processing $40$B tokens, including 2B from MC data.
% We sample $10$BT English data from \textit{SlimPajama}~\citep{cerebras2023slimpajama}, prompt \textit{Mistral-7B-Instruct-v0.1} with: "\textit{Write multiple-choice questions and answers based on the document: [doc]}", and then obtain $8$B tokens of MC synthetic data. We upsample and integrate this into \thedata{} alongside general web data, ensuring MC data comprises 5\% of the total corpus. Given the improved performance of \themodelweb{}, we continue pretraining it for $9,000$ steps, processing $40$B tokens, including $2$B tokens from MC data. 
% We refer to this model as \themodelmc{}, as detailed in Table~\ref{tab:data-impact-model-details}.
% Specifically, we randomly sample $10$B English tokens from \textit{SlimPajama}~\citep{cerebras2023slimpajama}, and prompt \textit{Mistral-7B-Instruct-v0.1} using the instruction template, "\textit{Write multiple-choice questions and answers based on the document: [doc]}", to generate rephrased MC synthetic data. After processing, we obtain 8B tokens, which we upsample and integrate into \thedata{} alongside real web data, ensuring MC data comprises 5\% of the final training corpus. Given the improved performance of \themodelweb{}, we continue pretraining it for $9,000$ steps, processing $40$B tokens, with $2$B tokens from the MC data. We refer to this model as \themodelmc{}, as detailed in Table~\ref{tab:data-impact-model-details}.

Prior works~\cite{faysse2024croissantllm,zhang2024tinyllama,martins2024eurollm} highlights the importance of a cooldown phase for enhancing model capabilities. While \thedata{} emphasizes educational content, it lacks code and instruction data, such as question-answering (QA), compared to other LLMs. To address this, we introduce \textit{\textbf{cooldown data}} during this phase: Python-Edu~\cite{benallal2024smollmcorpus}, an educational Python dataset from The Stack (4B tokens), and WebInstruct~\citep{yue2024mammoth2}, a curated QA dataset (0.8B tokens) from the web. They are up-sampled and mixed with the previous-stage data (Table~\ref{tab:data-impact-model-details}), forming 30\% of the total. The model undergoes an additional $20$B-token training phase using a reduced learning rate\footnote{We apply a constant learning rate schedule: $6\times 10^{-4}$ for earlier pretraining phases and $6\times 10^{-5}$ for cooldown.}. Notably, cooldown data constitutes less than $6$B tokens, accounting for only $0.3\%$ of total training tokens. We denote this final cooldown-trained model as \textbf{\themodelcool}.
% We upsample the cooldown data and mix it with the data from the previous stage, with cooldown data comprising 30\% of the total. The model is then trained on this new mix (Table~\ref{tab:data-impact-model-details}) for an additional $20$B tokens using a smaller learning rate.\footnote{We use a constant learning rate schedule: $6\times 10^{-4}$ for earlier pretraining phases and $6\times 10^{-5}$ for the cooldown stage.} Notably, cooldown data represents only a small fraction of the entire corpus, contributing fewer than 6B tokens, or approximately 0.3\% of total training tokens. We refer to this cooldown-trained model as \themodelcool.

We evaluate \themodelcool{} on all benchmarks in Sections~\ref{sec:benchmarks} and \ref{sec:real_web}, as well as Global-MMLU~\citep{singh2024global}, covering nine languages (excluding Welsh), in a $5$-shot setting. As shown in Table~\ref{tab:main_mmlu} for MMLU, \themodelcool{}, trained with additional rephrased synthetic and cooldown data, ranks among the top three models overall and achieves the highest performance in Swahili. Furthermore, Table~\ref{tab:all_understanding} in Appendix~\ref{sec:appendix-special-data} shows that \themodelcool{} surpasses \themodelweb{} across all languages for understanding and reasoning tasks, ranking as the top LLM on average. Remarkably, it is the \textit{\textbf{best-performing}} LLM for \textit{\textbf{Arabic, Italian, Indonesian, Swahili, and Welsh}}. Additionally, Tables~\ref{tab:french_ling_final} and~\ref{tab:local_culture_reasoning_final} in Appendix~\ref{sec:appendix-special-data} show that \themodelcool{}, despite being trained with limited additional special data, improves both French linguistic proficiency and Indonesian cultural reasoning. These findings highlight the effectiveness of rephrased synthetic and cooldown data in enhancing multilingual pretraining based on NLLB-translated data.

\section{Conclusion}
% We present \thedata{}, a multilingual dataset generated by machine-translating English source texts. Trained from scratch on this data, \themodel{} achieves competitive performance across nine non-English understanding and reasoning benchmarks, which matches or outperforms state-of-the-art multilingual LLMs trained using closed data, such as \textit{Llama3.2}, \textit{Qwen2.5}, and \textit{Gemma}. 
% We also show that  less than 5\% of TransWebEdu as
% domain-specific pretraining data sets a new
% state-of-the-art in Arabic, Italian, Indonesian,
% Swahili, and Welsh understanding and com-
% monsense reasoning tasks. In summary, we have shown that  constructing documents using a sentence-level translation model can yield robust performance for mutlilingual LLMs.
% Our approach provides a scalable solution for creating multilingual pretraining data, particularly for medium- and low-resource languages, contributing to advancements in multilingual NLP research.
We introduce \thedata{}, a multilingual dataset generated through machine translation of a high-quality English source dataset. Our model, \themodel{}, trained from scratch on this data, achieves competitive performance across nine non-English understanding and reasoning benchmarks, matching or surpassing state-of-the-art multilingual LLMs trained on closed data, such as \textit{Llama3.2}, \textit{Qwen2.5}, and \textit{Gemma} on average. Furthermore, we demonstrate that incorporating less than 5\% of TransWebEdu as domain-specific pretraining data establishes new state-of-the-art results in Arabic, Italian, Indonesian, Swahili, and Welsh for understanding and commonsense reasoning tasks. These findings highlight that constructing multilingual pretraining corpora using sentence-level translations can yield robust performance in multilingual LLMs. Our approach provides a scalable and efficient solution for creating multilingual pretraining data, particularly for medium- and low-resource languages, contributing to advancements in multilingual NLP research.

\section*{Limitations}
Our study yields promising results while also identifying areas for future exploration. 

\themodel{}, trained on \thedata{}, achieves significant performance gains across 10 multilingual benchmarks. Further improvements are observed with the addition of general web data, rephrased synthetic data, and code and web-instruct data. However, due to computational constraints, we did not conduct ablation studies to determine the optimal data mixing ratios beyond pretraining with \thedata{}. Future work will extend the ``Beyond Pretraining with Translation Data'' experiments in Section~\ref{sec:beyond} to explore optimal data integration strategies from diverse sources for \thedata{}.

In addition, our experiments focus on \themodel{}, a $1.3$B-parameter model that has shown promising results at this scale. However, it remains unclear whether the benefits of our translated pretraining data would persist or amplify in substantially larger models (e.g., $70$B+ parameters). Scaling up could provide deeper insights into multilingual learning dynamics and data efficiency. Future research will explore these aspects to validate and enhance the scalability of our multilingual pretraining approach. 

% We evaluate \themodelmc{} on the Global MMLU~\citep{singh2024global}, which covers 9 of our targeted languages. The comparison between TransWebLLM-web and TransWebLLM-mcSyn in Table~\ref{tab:data-impact-model-details} shows that adding a small amount of English MC synthetic data (only 2B trained tokens) yields the largest improvements in languages related to English—notably French, German, Italian, and Spanish. In contrast, more distant languages like Arabic, Swahili, and Russian show smaller gains, aligning with expected cross-lingual transfer patterns.

\bibliography{custom}

\begin{thebibliography}{71}
\providecommand{\natexlab}[1]{#1}

\bibitem[{Abdin et~al.(2024)Abdin, Aneja, Awadalla, Awadallah, Awan, Bach, Bahree, Bakhtiari, Bao, Behl et~al.}]{abdin2024phi}
Marah Abdin, Jyoti Aneja, Hany Awadalla, Ahmed Awadallah, Ammar~Ahmad Awan, Nguyen Bach, Amit Bahree, Arash Bakhtiari, Jianmin Bao, Harkirat Behl, et~al. 2024.
\newblock Phi-3 technical report: A highly capable language model locally on your phone.
\newblock \emph{arXiv preprint arXiv:2404.14219}.

\bibitem[{Alves et~al.(2024)Alves, Pombal, Guerreiro, Martins, Alves, Farajian, Peters, Rei, Fernandes, Agrawal et~al.}]{alves2024tower}
Duarte~M Alves, Jos{\'e} Pombal, Nuno~M Guerreiro, Pedro~H Martins, Jo{\~a}o Alves, Amin Farajian, Ben Peters, Ricardo Rei, Patrick Fernandes, Sweta Agrawal, et~al. 2024.
\newblock Tower: An open multilingual large language model for translation-related tasks.
\newblock \emph{arXiv preprint arXiv:2402.17733}.

\bibitem[{Bayes et~al.(2024)Bayes, Azime, Alabi, Kgomo, Eloundou, Proehl, Chen, Khadir, Etori, Muhammad et~al.}]{bayes2024uhura}
Edward Bayes, Israel~Abebe Azime, Jesujoba~O Alabi, Jonas Kgomo, Tyna Eloundou, Elizabeth Proehl, Kai Chen, Imaan Khadir, Naome~A Etori, Shamsuddeen~Hassan Muhammad, et~al. 2024.
\newblock Uhura: A benchmark for evaluating scientific question answering and truthfulness in low-resource african languages.
\newblock \emph{arXiv preprint arXiv:2412.00948}.

\bibitem[{Ben~Allal et~al.(2024)Ben~Allal, Lozhkov, Penedo, Wolf, and von Werra}]{benallal2024smollmcorpus}
Loubna Ben~Allal, Anton Lozhkov, Guilherme Penedo, Thomas Wolf, and Leandro von Werra. 2024.
\newblock \href {https://huggingface.co/datasets/HuggingFaceTB/smollm-corpus} {Smollm-corpus}.

\bibitem[{Bender et~al.(2021)Bender, Gebru, McMillan-Major, and Shmitchell}]{bender2021dangers}
Emily~M Bender, Timnit Gebru, Angelina McMillan-Major, and Shmargaret Shmitchell. 2021.
\newblock On the dangers of stochastic parrots: Can language models be too big?
\newblock In \emph{Proceedings of the 2021 ACM conference on fairness, accountability, and transparency}, pages 610--623.

\bibitem[{Biderman et~al.(2023)Biderman, Schoelkopf, Anthony, Bradley, O’Brien, Hallahan, Khan, Purohit, Prashanth, Raff et~al.}]{biderman2023pythia}
Stella Biderman, Hailey Schoelkopf, Quentin~Gregory Anthony, Herbie Bradley, Kyle O’Brien, Eric Hallahan, Mohammad~Aflah Khan, Shivanshu Purohit, USVSN~Sai Prashanth, Edward Raff, et~al. 2023.
\newblock Pythia: A suite for analyzing large language models across training and scaling.
\newblock In \emph{International Conference on Machine Learning}, pages 2397--2430. PMLR.

\bibitem[{Bird et~al.(2009)Bird, Klein, and Loper}]{bird2009natural}
Steven Bird, Ewan Klein, and Edward Loper. 2009.
\newblock \emph{Natural language processing with Python: analyzing text with the natural language toolkit}.
\newblock " O'Reilly Media, Inc.".

\bibitem[{Bisk et~al.(2020)Bisk, Zellers, Bras, Gao, and Choi}]{Bisk2020}
Yonatan Bisk, Rowan Zellers, Ronan~Le Bras, Jianfeng Gao, and Yejin Choi. 2020.
\newblock Piqa: Reasoning about physical commonsense in natural language.
\newblock In \emph{Thirty-Fourth AAAI Conference on Artificial Intelligence}.

\bibitem[{Broder(1997)}]{broder1997resemblance}
Andrei~Z Broder. 1997.
\newblock On the resemblance and containment of documents.
\newblock In \emph{Proceedings. Compression and Complexity of SEQUENCES 1997 (Cat. No. 97TB100171)}, pages 21--29. IEEE.

\bibitem[{Clark et~al.(2018)Clark, Cowhey, Etzioni, Khot, Sabharwal, Schoenick, and Tafjord}]{clark2018think}
Peter Clark, Isaac Cowhey, Oren Etzioni, Tushar Khot, Ashish Sabharwal, Carissa Schoenick, and Oyvind Tafjord. 2018.
\newblock Think you have solved question answering? try arc, the ai2 reasoning challenge.
\newblock \emph{arXiv preprint arXiv:1803.05457}.

\bibitem[{Conneau et~al.(2018)Conneau, Rinott, Lample, Williams, Bowman, Schwenk, and Stoyanov}]{conneau2018xnli}
Alexis Conneau, Ruty Rinott, Guillaume Lample, Adina Williams, Samuel~R. Bowman, Holger Schwenk, and Veselin Stoyanov. 2018.
\newblock Xnli: Evaluating cross-lingual sentence representations.
\newblock In \emph{Proceedings of the 2018 Conference on Empirical Methods in Natural Language Processing}. Association for Computational Linguistics.

\bibitem[{Costa-juss{\`a} et~al.(2022)Costa-juss{\`a}, Cross, {\c{C}}elebi, Elbayad, Heafield, Heffernan, Kalbassi, Lam, Licht, Maillard et~al.}]{costa2022no}
Marta~R Costa-juss{\`a}, James Cross, Onur {\c{C}}elebi, Maha Elbayad, Kenneth Heafield, Kevin Heffernan, Elahe Kalbassi, Janice Lam, Daniel Licht, Jean Maillard, et~al. 2022.
\newblock No language left behind: Scaling human-centered machine translation.
\newblock \emph{arXiv preprint arXiv:2207.04672}.

\bibitem[{Dao(2023)}]{dao2023flashattention}
Tri Dao. 2023.
\newblock Flashattention-2: Faster attention with better parallelism and work partitioning.
\newblock \emph{arXiv preprint arXiv:2307.08691}.

\bibitem[{Doshi et~al.(2024)Doshi, Dabre, and Bhattacharyya}]{doshi2024translationese}
Meet Doshi, Raj Dabre, and Pushpak Bhattacharyya. 2024.
\newblock Pretraining language models using translationese.
\newblock In \emph{Proceedings of the 2024 Conference on Empirical Methods in Natural Language Processing}, pages 5843--5862.

\bibitem[{Dou et~al.(2024{\natexlab{a}})Dou, Liu, Zeng, Guo, Zhou, Lu, and Lin}]{dou2024sailor}
Longxu Dou, Qian Liu, Guangtao Zeng, Jia Guo, Jiahui Zhou, Wei Lu, and Min Lin. 2024{\natexlab{a}}.
\newblock Sailor: Open language models for south-east asia.
\newblock \emph{arXiv preprint arXiv:2404.03608}.

\bibitem[{Dou et~al.(2024{\natexlab{b}})Dou, Liu, Zhou, Chen, Wang, Jin, Liu, Zhu, Du, Yang, Wang, Liu, Zhao, Feng, Mao, Yeung, Pipatanakul, Koto, Thu, Kydl{\'\i}{\v{c}}ek, Liu, Lin, Sripaisarnmongkol, Sae-Khow, Thongchim, Konkaew, Borijindargoon, Dao, Maneegard, Artkaew, Yong, Nguyen, Phatthiyaphaibun, Tran, Zhang, Chen, Pang, Du, Wan, Lu, and Lin}]{sailor2report}
Longxu Dou, Qian Liu, Fan Zhou, Changyu Chen, Zili Wang, Ziqi Jin, Zichen Liu, Tongyao Zhu, Cunxiao Du, Penghui Yang, Haonan Wang, Jiaheng Liu, Yongchi Zhao, Xiachong Feng, Xin Mao, Man~Tsung Yeung, Kunat Pipatanakul, Fajri Koto, Min~Si Thu, Hynek Kydl{\'\i}{\v{c}}ek, Zeyi Liu, Qunshu Lin, Sittipong Sripaisarnmongkol, Kridtaphad Sae-Khow, Nirattisai Thongchim, Taechawat Konkaew, Narong Borijindargoon, Anh Dao, Matichon Maneegard, Phakphum Artkaew, Zheng-Xin Yong, Quan Nguyen, Wannaphong Phatthiyaphaibun, Hoang~H. Tran, Mike Zhang, Shiqi Chen, Tianyu Pang, Chao Du, Xinyi Wan, Wei Lu, and Min Lin. 2024{\natexlab{b}}.
\newblock Sailor2: Sailing in south-east asia with inclusive multilingual llm.

\bibitem[{Dubey et~al.(2024)Dubey, Jauhri, Pandey, Kadian, Al-Dahle, Letman, Mathur, Schelten, Yang, Fan et~al.}]{dubey2024llama}
Abhimanyu Dubey, Abhinav Jauhri, Abhinav Pandey, Abhishek Kadian, Ahmad Al-Dahle, Aiesha Letman, Akhil Mathur, Alan Schelten, Amy Yang, Angela Fan, et~al. 2024.
\newblock The llama 3 herd of models.
\newblock \emph{arXiv preprint arXiv:2407.21783}.

\bibitem[{Ezeani et~al.(2019)Ezeani, Piao, Neale, Rayson, and Knight}]{ezeani-etal-2019-leveraging}
Ignatius Ezeani, Scott Piao, Steven Neale, Paul Rayson, and Dawn Knight. 2019.
\newblock \href {https://doi.org/10.18653/v1/W19-4332} {Leveraging pre-trained embeddings for {W}elsh taggers}.
\newblock In \emph{Proceedings of the 4th Workshop on Representation Learning for NLP (RepL4NLP-2019)}, pages 270--280, Florence, Italy. Association for Computational Linguistics.

\bibitem[{Faysse et~al.(2024)Faysse, Fernandes, Guerreiro, Loison, Alves, Corro, Boizard, Alves, Rei, Martins et~al.}]{faysse2024croissantllm}
Manuel Faysse, Patrick Fernandes, Nuno Guerreiro, Ant{\'o}nio Loison, Duarte Alves, Caio Corro, Nicolas Boizard, Jo{\~a}o Alves, Ricardo Rei, Pedro Martins, et~al. 2024.
\newblock Croissantllm: A truly bilingual french-english language model.
\newblock \emph{arXiv preprint arXiv:2402.00786}.

\bibitem[{Gao et~al.(2020)Gao, Biderman, Black, Golding, Hoppe, Foster, Phang, He, Thite, Nabeshima et~al.}]{gao2020pile}
Leo Gao, Stella Biderman, Sid Black, Laurence Golding, Travis Hoppe, Charles Foster, Jason Phang, Horace He, Anish Thite, Noa Nabeshima, et~al. 2020.
\newblock The pile: An 800gb dataset of diverse text for language modeling.
\newblock \emph{arXiv preprint arXiv:2101.00027}.

\bibitem[{Hendrycks et~al.(2021)Hendrycks, Burns, Basart, Zou, Mazeika, Song, and Steinhardt}]{hendrycks2021measuring}
Dan Hendrycks, Collin Burns, Steven Basart, Andy Zou, Mantas Mazeika, Dawn Song, and Jacob Steinhardt. 2021.
\newblock Measuring massive multitask language understanding.
\newblock \emph{arXiv preprint arXiv:2009.03300}.

\bibitem[{Imani et~al.(2023)Imani, Lin, Kargaran, Severini, Jalili~Sabet, Kassner, Ma, Schmid, Martins, Yvon, and Sch{\"u}tze}]{imanigooghari-etal-2023-glot500}
Ayyoob Imani, Peiqin Lin, Amir~Hossein Kargaran, Silvia Severini, Masoud Jalili~Sabet, Nora Kassner, Chunlan Ma, Helmut Schmid, Andr{\'e} Martins, Fran{\c{c}}ois Yvon, and Hinrich Sch{\"u}tze. 2023.
\newblock \href {https://doi.org/10.18653/v1/2023.acl-long.61} {Glot500: Scaling multilingual corpora and language models to 500 languages}.
\newblock In \emph{Proceedings of the 61st Annual Meeting of the Association for Computational Linguistics (Volume 1: Long Papers)}, pages 1082--1117, Toronto, Canada. Association for Computational Linguistics.

\bibitem[{Johannes~Welbl(2017)}]{SciQ}
Matt~Gardner Johannes~Welbl, Nelson F.~Liu. 2017.
\newblock Crowdsourcing multiple choice science questions.

\bibitem[{Joshi et~al.(2020)Joshi, Santy, Budhiraja, Bali, and Choudhury}]{joshi2020state}
Pratik Joshi, Sebastin Santy, Amar Budhiraja, Kalika Bali, and Monojit Choudhury. 2020.
\newblock The state and fate of linguistic diversity and inclusion in the nlp world.
\newblock \emph{arXiv preprint arXiv:2004.09095}.

\bibitem[{Karpathy(2022)}]{Karpathy2022}
Andrej Karpathy. 2022.
\newblock \text{NanoGPT}.
\newblock \url{https://github.com/karpathy/nanoGPT}.

\bibitem[{Karpathy(2024)}]{Karpathy2024}
Andrej Karpathy. 2024.
\newblock \text{llm.c}.
\newblock \url{https://github.com/karpathy/llm.c }.

\bibitem[{Kocmi et~al.(2024)Kocmi, Avramidis, Bawden, Bojar, Dvorkovich, Federmann, Fishel, Freitag, Gowda, Grundkiewicz et~al.}]{kocmi2024preliminary}
Tom Kocmi, Eleftherios Avramidis, Rachel Bawden, Ondrej Bojar, Anton Dvorkovich, Christian Federmann, Mark Fishel, Markus Freitag, Thamme Gowda, Roman Grundkiewicz, et~al. 2024.
\newblock Preliminary wmt24 ranking of general mt systems and llms.
\newblock \emph{arXiv preprint arXiv:2407.19884}.

\bibitem[{Kreutzer et~al.(2022)Kreutzer, Caswell, Wang, Wahab, van Esch, Ulzii-Orshikh, Tapo, Subramani, Sokolov, Sikasote et~al.}]{kreutzer2022quality}
Julia Kreutzer, Isaac Caswell, Lisa Wang, Ahsan Wahab, Daan van Esch, Nasanbayar Ulzii-Orshikh, Allahsera Tapo, Nishant Subramani, Artem Sokolov, Claytone Sikasote, et~al. 2022.
\newblock Quality at a glance: An audit of web-crawled multilingual datasets.
\newblock \emph{Transactions of the Association for Computational Linguistics}, 10:50--72.

\bibitem[{Kudugunta et~al.(2024)Kudugunta, Caswell, Zhang, Garcia, Xin, Kusupati, Stella, Bapna, and Firat}]{kudugunta2024madlad}
Sneha Kudugunta, Isaac Caswell, Biao Zhang, Xavier Garcia, Derrick Xin, Aditya Kusupati, Romi Stella, Ankur Bapna, and Orhan Firat. 2024.
\newblock Madlad-400: A multilingual and document-level large audited dataset.
\newblock \emph{Advances in Neural Information Processing Systems}, 36.

\bibitem[{Kwon et~al.(2023)Kwon, Li, Zhuang, Sheng, Zheng, Yu, Gonzalez, Zhang, and Stoica}]{kwon2023efficient}
Woosuk Kwon, Zhuohan Li, Siyuan Zhuang, Ying Sheng, Lianmin Zheng, Cody~Hao Yu, Joseph~E. Gonzalez, Hao Zhang, and Ion Stoica. 2023.
\newblock Efficient memory management for large language model serving with pagedattention.
\newblock In \emph{Proceedings of the ACM SIGOPS 29th Symposium on Operating Systems Principles}.

\bibitem[{Lai et~al.(2023)Lai, Van~Nguyen, Ngo, Nguyen, Dernoncourt, Rossi, and Nguyen}]{lai2023okapi}
Viet~Dac Lai, Chien Van~Nguyen, Nghia~Trung Ngo, Thuat Nguyen, Franck Dernoncourt, Ryan~A Rossi, and Thien~Huu Nguyen. 2023.
\newblock Okapi: Instruction-tuned large language models in multiple languages with reinforcement learning from human feedback.
\newblock \emph{arXiv preprint arXiv:2307.16039}.

\bibitem[{Lee et~al.(2021)Lee, Ippolito, Nystrom, Zhang, Eck, Callison-Burch, and Carlini}]{lee2021deduplicating}
Katherine Lee, Daphne Ippolito, Andrew Nystrom, Chiyuan Zhang, Douglas Eck, Chris Callison-Burch, and Nicholas Carlini. 2021.
\newblock Deduplicating training data makes language models better.
\newblock \emph{arXiv preprint arXiv:2107.06499}.

\bibitem[{Li et~al.(2023)Li, Allal, Zi, Muennighoff, Kocetkov, Mou, Marone, Akiki, Li, Chim et~al.}]{li2023starcoder}
Raymond Li, Loubna~Ben Allal, Yangtian Zi, Niklas Muennighoff, Denis Kocetkov, Chenghao Mou, Marc Marone, Christopher Akiki, Jia Li, Jenny Chim, et~al. 2023.
\newblock Starcoder: may the source be with you!
\newblock \emph{arXiv preprint arXiv:2305.06161}.

\bibitem[{Lin et~al.(2021{\natexlab{a}})Lin, Hilton, and Evans}]{lin2021truthfulqa}
Stephanie Lin, Jacob Hilton, and Owain Evans. 2021{\natexlab{a}}.
\newblock Truthfulqa: Measuring how models mimic human falsehoods.
\newblock \emph{arXiv preprint arXiv:2109.07958}.

\bibitem[{Lin et~al.(2021{\natexlab{b}})Lin, Mihaylov, Artetxe, Wang, Chen, Simig, Ott, Goyal, Bhosale, Du et~al.}]{lin2021few}
Xi~Victoria Lin, Todor Mihaylov, Mikel Artetxe, Tianlu Wang, Shuohui Chen, Daniel Simig, Myle Ott, Naman Goyal, Shruti Bhosale, Jingfei Du, et~al. 2021{\natexlab{b}}.
\newblock Few-shot learning with multilingual language models.
\newblock \emph{arXiv preprint arXiv:2112.10668}.

\bibitem[{Lozhkov et~al.(2024)Lozhkov, Ben~Allal, von Werra, and Wolf}]{lozhkov2024fineweb-edu}
Anton Lozhkov, Loubna Ben~Allal, Leandro von Werra, and Thomas Wolf. 2024.
\newblock \href {https://doi.org/10.57967/hf/2497} {Fineweb-edu}.

\bibitem[{Maini et~al.(2024)Maini, Seto, Bai, Grangier, Zhang, and Jaitly}]{maini2024rephrasing}
Pratyush Maini, Skyler Seto, He~Bai, David Grangier, Yizhe Zhang, and Navdeep Jaitly. 2024.
\newblock Rephrasing the web: A recipe for compute and data-efficient language modeling.
\newblock \emph{arXiv preprint arXiv:2401.16380}.

\bibitem[{Martins et~al.(2024)Martins, Fernandes, Alves, Guerreiro, Rei, Alves, Pombal, Farajian, Faysse, Klimaszewski et~al.}]{martins2024eurollm}
Pedro~Henrique Martins, Patrick Fernandes, Jo{\~a}o Alves, Nuno~M Guerreiro, Ricardo Rei, Duarte~M Alves, Jos{\'e} Pombal, Amin Farajian, Manuel Faysse, Mateusz Klimaszewski, et~al. 2024.
\newblock Eurollm: Multilingual language models for europe.
\newblock \emph{arXiv preprint arXiv:2409.16235}.

\bibitem[{McIntosh-Smith et~al.(2024)McIntosh-Smith, Alam, and Woods}]{mcintosh2024isambard}
Simon McIntosh-Smith, Sadaf~R Alam, and Christopher Woods. 2024.
\newblock Isambard-ai: a leadership class supercomputer optimised specifically for artificial intelligence.
\newblock \emph{arXiv preprint arXiv:2410.11199}.

\bibitem[{Moslem et~al.(2023)Moslem, Haque, and Way}]{moslem2023fine}
Yasmin Moslem, Rejwanul Haque, and Andy Way. 2023.
\newblock Fine-tuning large language models for adaptive machine translation.
\newblock \emph{arXiv preprint arXiv:2312.12740}.

\bibitem[{Mostafazadeh et~al.(2016)Mostafazadeh, Chambers, He, Parikh, Batra, Vanderwende, Kohli, and Allen}]{mostafazadeh2016corpus}
Nasrin Mostafazadeh, Nathanael Chambers, Xiaodong He, Devi Parikh, Dhruv Batra, Lucy Vanderwende, Pushmeet Kohli, and James Allen. 2016.
\newblock A corpus and evaluation framework for deeper understanding of commonsense stories.
\newblock \emph{arXiv preprint arXiv:1604.01696}.

\bibitem[{Muennighoff et~al.(2024)Muennighoff, Rush, Barak, Le~Scao, Tazi, Piktus, Pyysalo, Wolf, and Raffel}]{muennighoff2024datascaling}
Niklas Muennighoff, Alexander Rush, Boaz Barak, Teven Le~Scao, Nouamane Tazi, Aleksandra Piktus, Sampo Pyysalo, Thomas Wolf, and Colin~A Raffel. 2024.
\newblock Scaling data-constrained language models.
\newblock \emph{Advances in Neural Information Processing Systems}, 36.

\bibitem[{Oladipo et~al.(2023)Oladipo, Adeyemi, Ahia, Owodunni, Ogundepo, Adelani, and Lin}]{oladipo2023better}
Akintunde Oladipo, Mofetoluwa Adeyemi, Orevaoghene Ahia, Abraham Owodunni, Odunayo Ogundepo, David Adelani, and Jimmy Lin. 2023.
\newblock Better quality pre-training data and t5 models for african languages.
\newblock In \emph{Proceedings of the 2023 Conference on Empirical Methods in Natural Language Processing}, pages 158--168.

\bibitem[{{Ortiz Su{'a}rez} et~al.(2019){Ortiz Su{'a}rez}, Sagot, and Romary}]{OrtizSuarezSagotRomary2019}
Pedro~Javier {Ortiz Su{'a}rez}, Benoit Sagot, and Laurent Romary. 2019.
\newblock \href {https://doi.org/10.14618/ids-pub-9021} {Asynchronous pipelines for processing huge corpora on medium to low resource infrastructures}.
\newblock Proceedings of the Workshop on Challenges in the Management of Large Corpora (CMLC-7) 2019. Cardiff, 22nd July 2019, pages 9 -- 16, Mannheim. Leibniz-Institut f{"u}r Deutsche Sprache.

\bibitem[{Penedo et~al.(2024)Penedo, Kydlíček, allal, Lozhkov, Mitchell, Raffel, Werra, and Wolf}]{penedo2024finewebdatasetsdecantingweb}
Guilherme Penedo, Hynek Kydlíček, Loubna~Ben allal, Anton Lozhkov, Margaret Mitchell, Colin Raffel, Leandro~Von Werra, and Thomas Wolf. 2024.
\newblock \href {https://arxiv.org/abs/2406.17557} {The fineweb datasets: Decanting the web for the finest text data at scale}.
\newblock \emph{Preprint}, arXiv:2406.17557.

\bibitem[{Ponti et~al.(2020)Ponti, Glava{\v{s}}, Majewska, Liu, Vuli{\'c}, and Korhonen}]{ponti2020xcopa}
Edoardo~Maria Ponti, Goran Glava{\v{s}}, Olga Majewska, Qianchu Liu, Ivan Vuli{\'c}, and Anna Korhonen. 2020.
\newblock Xcopa: A multilingual dataset for causal commonsense reasoning.
\newblock \emph{arXiv preprint arXiv:2005.00333}.

\bibitem[{Rae et~al.(2021)Rae, Borgeaud, Cai, Millican, Hoffmann, Song, Aslanides, Henderson, Ring, Young et~al.}]{rae2021scaling}
Jack~W Rae, Sebastian Borgeaud, Trevor Cai, Katie Millican, Jordan Hoffmann, Francis Song, John Aslanides, Sarah Henderson, Roman Ring, Susannah Young, et~al. 2021.
\newblock Scaling language models: Methods, analysis \& insights from training gopher.
\newblock \emph{arXiv preprint arXiv:2112.11446}.

\bibitem[{Roemmele et~al.(2011)Roemmele, Bejan, and Gordon}]{roemmele2011choice}
Melissa Roemmele, Cosmin~Adrian Bejan, and Andrew~S Gordon. 2011.
\newblock Choice of plausible alternatives: An evaluation of commonsense causal reasoning.
\newblock In \emph{2011 AAAI spring symposium series}.

\bibitem[{Sengupta et~al.(2023)Sengupta, Sahu, Jia, Katipomu, Li, Koto, Marshall, Gosal, Liu, Chen, Afzal, Kamboj, Pandit, Pal, Pradhan, Mujahid, Baali, Han, Bsharat, Aji, Shen, Liu, Vassilieva, Hestness, Hock, Feldman, Lee, Jackson, Ren, Nakov, Baldwin, and Xing}]{sengupta2023jais}
Neha Sengupta, Sunil~Kumar Sahu, Bokang Jia, Satheesh Katipomu, Haonan Li, Fajri Koto, William Marshall, Gurpreet Gosal, Cynthia Liu, Zhiming Chen, Osama~Mohammed Afzal, Samta Kamboj, Onkar Pandit, Rahul Pal, Lalit Pradhan, Zain~Muhammad Mujahid, Massa Baali, Xudong Han, Sondos~Mahmoud Bsharat, Alham~Fikri Aji, Zhiqiang Shen, Zhengzhong Liu, Natalia Vassilieva, Joel Hestness, Andy Hock, Andrew Feldman, Jonathan Lee, Andrew Jackson, Hector~Xuguang Ren, Preslav Nakov, Timothy Baldwin, and Eric Xing. 2023.
\newblock \href {https://arxiv.org/abs/2308.16149} {Jais and jais-chat: Arabic-centric foundation and instruction-tuned open generative large language models}.
\newblock \emph{Preprint}, arXiv:2308.16149.

\bibitem[{Shliazhko et~al.(2022)Shliazhko, Fenogenova, Tikhonova, Mikhailov, Kozlova, and Shavrina}]{shliazhko2022mgpt}
Oleh Shliazhko, Alena Fenogenova, Maria Tikhonova, Vladislav Mikhailov, Anastasia Kozlova, and Tatiana Shavrina. 2022.
\newblock mgpt: Few-shot learners go multilingual.
\newblock \emph{arXiv preprint arXiv:2204.07580}.

\bibitem[{Shoeybi et~al.(2019)Shoeybi, Patwary, Puri, LeGresley, Casper, and Catanzaro}]{shoeybi2019megatron}
Mohammad Shoeybi, Mostofa Patwary, Raul Puri, Patrick LeGresley, Jared Casper, and Bryan Catanzaro. 2019.
\newblock Megatron-lm: Training multi-billion parameter language models using model parallelism.
\newblock \emph{arXiv preprint arXiv:1909.08053}.

\bibitem[{Singh et~al.(2024)Singh, Romanou, Fourrier, Adelani, Ngui, Vila-Suero, Limkonchotiwat, Marchisio, Leong, Susanto et~al.}]{singh2024global}
Shivalika Singh, Angelika Romanou, Cl{\'e}mentine Fourrier, David~I Adelani, Jian~Gang Ngui, Daniel Vila-Suero, Peerat Limkonchotiwat, Kelly Marchisio, Wei~Qi Leong, Yosephine Susanto, et~al. 2024.
\newblock Global mmlu: Understanding and addressing cultural and linguistic biases in multilingual evaluation.
\newblock \emph{arXiv preprint arXiv:2412.03304}.

\bibitem[{Soboleva et~al.(2023)Soboleva, Al-Khateeb, Myers, Steeves, Hestness, and Dey}]{cerebras2023slimpajama}
Daria Soboleva, Faisal Al-Khateeb, Robert Myers, Jacob~R Steeves, Joel Hestness, and Nolan Dey. 2023.
\newblock \href {https://huggingface.co/datasets/cerebras/SlimPajama-627B} {{SlimPajama: A 627B token cleaned and deduplicated version of RedPajama}}.

\bibitem[{Stahlberg(2020)}]{stahlberg2020neural}
Felix Stahlberg. 2020.
\newblock Neural machine translation: A review.
\newblock \emph{Journal of Artificial Intelligence Research}, 69:343--418.

\bibitem[{Team et~al.(2024)Team, Riviere, Pathak, Sessa, Hardin, Bhupatiraju, Hussenot, Mesnard, Shahriari, Ram{\'e} et~al.}]{team2024gemma}
Gemma Team, Morgane Riviere, Shreya Pathak, Pier~Giuseppe Sessa, Cassidy Hardin, Surya Bhupatiraju, L{\'e}onard Hussenot, Thomas Mesnard, Bobak Shahriari, Alexandre Ram{\'e}, et~al. 2024.
\newblock Gemma 2: Improving open language models at a practical size.
\newblock \emph{arXiv preprint arXiv:2408.00118}.

\bibitem[{Tikhonov and Ryabinin(2021)}]{tikhonov2021s}
Alexey Tikhonov and Max Ryabinin. 2021.
\newblock It's all in the heads: Using attention heads as a baseline for cross-lingual transfer in commonsense reasoning.
\newblock \emph{arXiv preprint arXiv:2106.12066}.

\bibitem[{Touvron et~al.(2023)Touvron, Martin, Stone, Albert, Almahairi, Babaei, Bashlykov, Batra, Bhargava, Bhosale et~al.}]{touvron2023llama2}
Hugo Touvron, Louis Martin, Kevin Stone, Peter Albert, Amjad Almahairi, Yasmine Babaei, Nikolay Bashlykov, Soumya Batra, Prajjwal Bhargava, Shruti Bhosale, et~al. 2023.
\newblock Llama 2: Open foundation and fine-tuned chat models.
\newblock \emph{arXiv preprint arXiv:2307.09288}.

\bibitem[{Urbizu et~al.(2023)Urbizu, San~Vicente, Saralegi, and Corral}]{urbizu2023mtrescue}
Gorka Urbizu, I{\~n}aki San~Vicente, Xabier Saralegi, and Ander Corral. 2023.
\newblock Not enough data to pre-train your language model? mt to the rescue!
\newblock In \emph{Findings of the Association for Computational Linguistics: ACL 2023}, pages 3826--3836.

\bibitem[{Weber et~al.(2024)Weber, Fu, Anthony, Oren, Adams, Alexandrov, Lyu, Nguyen, Yao, Adams, Athiwaratkun, Chalamala, Chen, Ryabinin, Dao, Liang, R\'{e}, Rish, and Zhang}]{NEURIPS2024_d3449733}
Maurice Weber, Dan Fu, Quentin Anthony, Yonatan Oren, Shane Adams, Anton Alexandrov, Xiaozhong Lyu, Huu Nguyen, Xiaozhe Yao, Virginia Adams, Ben Athiwaratkun, Rahul Chalamala, Kezhen Chen, Max Ryabinin, Tri Dao, Percy~S Liang, Christopher R\'{e}, Irina Rish, and Ce~Zhang. 2024.
\newblock \href {https://proceedings.neurips.cc/paper_files/paper/2024/file/d34497330b1fd6530f7afd86d0df9f76-Paper-Datasets_and_Benchmarks_Track.pdf} {Redpajama: an open dataset for training large language models}.
\newblock In \emph{Advances in Neural Information Processing Systems}, volume~37, pages 116462--116492. Curran Associates, Inc.

\bibitem[{Wenzek et~al.(2020)Wenzek, Lachaux, Conneau, Chaudhary, Guzm{\'a}n, Joulin, and Grave}]{wenzek-etal-2020-ccnet}
Guillaume Wenzek, Marie-Anne Lachaux, Alexis Conneau, Vishrav Chaudhary, Francisco Guzm{\'a}n, Armand Joulin, and Edouard Grave. 2020.
\newblock \href {https://aclanthology.org/2020.lrec-1.494/} {{CCN}et: Extracting high quality monolingual datasets from web crawl data}.
\newblock In \emph{Proceedings of the Twelfth Language Resources and Evaluation Conference}, pages 4003--4012, Marseille, France. European Language Resources Association.

\bibitem[{Wibowo et~al.(2023)Wibowo, Fuadi, Nityasya, Prasojo, and Aji}]{wibowo2023copal}
Haryo~Akbarianto Wibowo, Erland~Hilman Fuadi, Made~Nindyatama Nityasya, Radityo~Eko Prasojo, and Alham~Fikri Aji. 2023.
\newblock Copal-id: Indonesian language reasoning with local culture and nuances.
\newblock \emph{arXiv preprint arXiv:2311.01012}.

\bibitem[{Williams et~al.(2018)Williams, Nangia, and Bowman}]{williams-etal-2018-broad}
Adina Williams, Nikita Nangia, and Samuel Bowman. 2018.
\newblock \href {https://doi.org/10.18653/v1/N18-1101} {A broad-coverage challenge corpus for sentence understanding through inference}.
\newblock In \emph{Proceedings of the 2018 Conference of the North {A}merican Chapter of the Association for Computational Linguistics: Human Language Technologies, Volume 1 (Long Papers)}, pages 1112--1122, New Orleans, Louisiana. Association for Computational Linguistics.

\bibitem[{Workshop et~al.(2022)Workshop, Scao, Fan, Akiki, Pavlick, Ili{\'c}, Hesslow, Castagn{\'e}, Luccioni, Yvon et~al.}]{le2023bloom}
BigScience Workshop, Teven~Le Scao, Angela Fan, Christopher Akiki, Ellie Pavlick, Suzana Ili{\'c}, Daniel Hesslow, Roman Castagn{\'e}, Alexandra~Sasha Luccioni, Fran{\c{c}}ois Yvon, et~al. 2022.
\newblock Bloom: A 176b-parameter open-access multilingual language model.
\newblock \emph{arXiv preprint arXiv:2211.05100}.

\bibitem[{Xue et~al.(2021)Xue, Constant, Roberts, Kale, Al-Rfou, Siddhant, Barua, and Raffel}]{xue-etal-2021-mt5}
Linting Xue, Noah Constant, Adam Roberts, Mihir Kale, Rami Al-Rfou, Aditya Siddhant, Anirudh Barua, and Colin Raffel. 2021.
\newblock {mT5}: A massively multilingual pre-trained text-to-text transformer.
\newblock \emph{arXiv preprint arXiv:2010.11934}.

\bibitem[{Yang et~al.(2024)Yang, Yang, Hui, Zheng, Yu, Zhou, Li, Li, Liu, Huang, Dong, Wei, Lin, Tang, Wang, Yang, Tu, Zhang, Ma, Yang, Xu, Zhou, Bai, He, Lin, Dang, Lu, Chen, Yang, Li, Xue, Ni, Zhang, Wang, Peng, Men, Gao, Lin, Wang, Bai, Tan, Zhu, Li, Liu, Ge, Deng, Zhou, Ren, Zhang, Wei, Ren, Liu, Fan, Yao, Zhang, Wan, Chu, Liu, Cui, Zhang, Guo, and Fan}]{yang2024qwen2technicalreport}
An~Yang, Baosong Yang, Binyuan Hui, Bo~Zheng, Bowen Yu, Chang Zhou, Chengpeng Li, Chengyuan Li, Dayiheng Liu, Fei Huang, Guanting Dong, Haoran Wei, Huan Lin, Jialong Tang, Jialin Wang, Jian Yang, Jianhong Tu, Jianwei Zhang, Jianxin Ma, Jianxin Yang, Jin Xu, Jingren Zhou, Jinze Bai, Jinzheng He, Junyang Lin, Kai Dang, Keming Lu, Keqin Chen, Kexin Yang, Mei Li, Mingfeng Xue, Na~Ni, Pei Zhang, Peng Wang, Ru~Peng, Rui Men, Ruize Gao, Runji Lin, Shijie Wang, Shuai Bai, Sinan Tan, Tianhang Zhu, Tianhao Li, Tianyu Liu, Wenbin Ge, Xiaodong Deng, Xiaohuan Zhou, Xingzhang Ren, Xinyu Zhang, Xipin Wei, Xuancheng Ren, Xuejing Liu, Yang Fan, Yang Yao, Yichang Zhang, Yu~Wan, Yunfei Chu, Yuqiong Liu, Zeyu Cui, Zhenru Zhang, Zhifang Guo, and Zhihao Fan. 2024.
\newblock \href {https://arxiv.org/abs/2407.10671} {Qwen2 technical report}.
\newblock \emph{Preprint}, arXiv:2407.10671.

\bibitem[{Yang et~al.(2025)Yang, Yang, Zhang, Hui, Zheng, Yu, Li, Liu, Huang, Wei, Lin, Yang, Tu, Zhang, Yang, Yang, Zhou, Lin, Dang, Lu, Bao, Yang, Yu, Li, Xue, Zhang, Zhu, Men, Lin, Li, Tang, Xia, Ren, Ren, Fan, Su, Zhang, Wan, Liu, Cui, Zhang, and Qiu}]{qwen2025qwen25technicalreport}
An~Yang, Baosong Yang, Beichen Zhang, Binyuan Hui, Bo~Zheng, Bowen Yu, Chengyuan Li, Dayiheng Liu, Fei Huang, Haoran Wei, Huan Lin, Jian Yang, Jianhong Tu, Jianwei Zhang, Jianxin Yang, Jiaxi Yang, Jingren Zhou, Junyang Lin, Kai Dang, Keming Lu, Keqin Bao, Kexin Yang, Le~Yu, Mei Li, Mingfeng Xue, Pei Zhang, Qin Zhu, Rui Men, Runji Lin, Tianhao Li, Tianyi Tang, Tingyu Xia, Xingzhang Ren, Xuancheng Ren, Yang Fan, Yang Su, Yichang Zhang, Yu~Wan, Yuqiong Liu, Zeyu Cui, Zhenru Zhang, and Zihan Qiu. 2025.
\newblock \href {https://arxiv.org/abs/2412.15115} {Qwen2.5 technical report}.
\newblock \emph{Preprint}, arXiv:2412.15115.

\bibitem[{Yang et~al.(2023)Yang, Chiang, Zheng, Gonzalez, and Stoica}]{yang2023rethinking}
Shuo Yang, Wei-Lin Chiang, Lianmin Zheng, Joseph~E Gonzalez, and Ion Stoica. 2023.
\newblock Rethinking benchmark and contamination for language models with rephrased samples.
\newblock \emph{arXiv preprint arXiv:2311.04850}.

\bibitem[{Yang et~al.(2019)Yang, Zhang, Tar, and Baldridge}]{yang-etal-2019-paws}
Yinfei Yang, Yuan Zhang, Chris Tar, and Jason Baldridge. 2019.
\newblock \href {https://doi.org/10.18653/v1/D19-1382} {{PAWS}-{X}: A cross-lingual adversarial dataset for paraphrase identification}.
\newblock In \emph{Proceedings of the 2019 Conference on Empirical Methods in Natural Language Processing and the 9th International Joint Conference on Natural Language Processing (EMNLP-IJCNLP)}, pages 3687--3692, Hong Kong, China. Association for Computational Linguistics.

\bibitem[{Yue et~al.(2024)Yue, Zheng, Zhang, and Chen}]{yue2024mammoth2}
Xiang Yue, Tuney Zheng, Ge~Zhang, and Wenhu Chen. 2024.
\newblock Mammoth2: Scaling instructions from the web.
\newblock \emph{arXiv preprint arXiv:2405.03548}.

\bibitem[{Zellers et~al.(2019)Zellers, Holtzman, Bisk, Farhadi, and Choi}]{zellers2019hellaswag}
Rowan Zellers, Ari Holtzman, Yonatan Bisk, Ali Farhadi, and Yejin Choi. 2019.
\newblock Hellaswag: Can a machine really finish your sentence?
\newblock In \emph{Proceedings of the 57th Annual Meeting of the Association for Computational Linguistics}.

\bibitem[{Zhang et~al.(2024)Zhang, Zeng, Wang, and Lu}]{zhang2024tinyllama}
Peiyuan Zhang, Guangtao Zeng, Tianduo Wang, and Wei Lu. 2024.
\newblock Tinyllama: An open-source small language model.
\newblock \emph{arXiv preprint arXiv:2401.02385}.

\end{thebibliography}
\appendix
\section*{Appendix}
This appendix provides additional technical details on our approach and supplementary evaluation results for the main paper.

\section{Translation Pipeline}
\begin{figure*}[!t]
\includegraphics[width=1.0\textwidth]{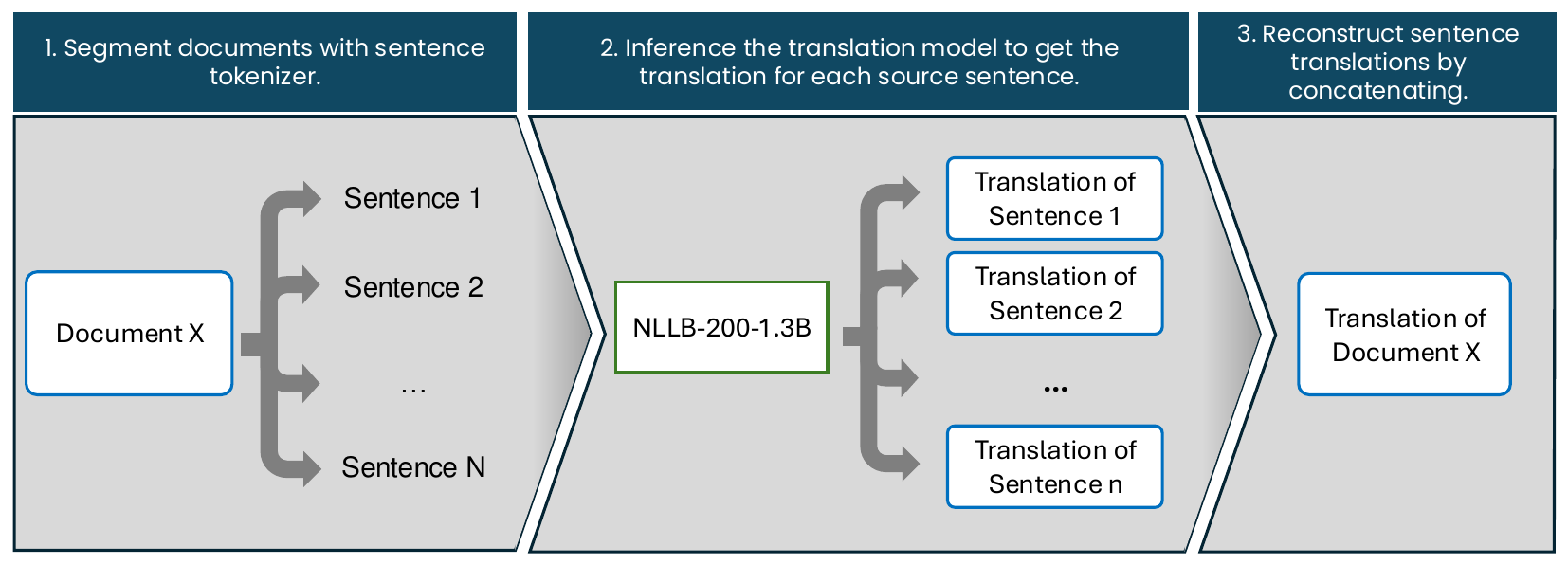}
\caption{Step-by-step illustration of the translation pipeline to obtain \thedata{}.}
\label{fig:trans_pip}
\end{figure*}

The detailed translation pipeline to produce \thedata{} is shown in Figure~\ref{fig:trans_pip}.
\label{sec:trans_pip}

\section{Data Statistics of \thedata}
\label{sec:appendix-nllb-statistics}
\begin{table}[!t]
\small
\centering
\begin{tabular}{l|cc}
\toprule
Language & \multicolumn{1}{c}{Tokens (B)} & \multicolumn{1}{c}{Avg. Doc Length (tokens)} \\
\midrule
Arabic	&	311.35	&	3201	\\
English	&	114.95	&	1182	\\
French	&	143.71	&	1479	\\
German	&	140.70	&	1447	\\
Indonesian	&	174.12	&	1792	\\
Italian	&	140.32	&	1447	\\
Russian	&	157.40	&	1618	\\
Spanish	&	140.99	&	1449	\\
Swahili	&	183.55	&	1887	\\
Welsh	&	201.49	&	2071	\\
\midrule
Total & 1708.58 & 1757 \\
\bottomrule
\end{tabular}
\caption{Statistics of \thedata Dataset, measured in Llama2 tokenizer.}
\label{tab:nllb_statistics}
\end{table}
The statistics of \thedata{} are shown in Table~\ref{tab:nllb_statistics}.

\section{Hyperparameters Settings of Model Pretrainning}
\label{sec:appendix-model-arch-params}
\begin{table}[!t]
% \begin{center}
\small 
\centering
\setlength{\tabcolsep}{8pt}
% \resizebox{\columnwidth}{!}{%
\begin{tabular}{l|c}
\toprule
 Hyperparameter &  Value\\
\midrule
Sequence Length &   2048 \\
    Number of Layers &    24 \\
   Embedding Size &    2048 \\
   FFN Hidden Size &   5504 \\
   Number of Heads &   16 \\
   Position Encodings &   RoPE \\
   Activation Function &   SwiGLU\\
   Layer Norm &   RMSNorm \\
   Learning Rate &   6E-4 \\
   Batch Size &   1024 \\
   Vocabulary Size &    32000 \\
 \midrule
   Embedding Parameters &   0.13B \\
   Non-Embedding Parameters &   1.21B\\
   Total Parameters &   1.34B\\
\bottomrule
\end{tabular}
% }
% \end{center}
\caption{Model and pretraining hyperparameters.} 
\label{tab:model-arch}
\end{table}
Pretraining hyperparameter settings are shown in Table~\ref{tab:model-arch}.

\section{Specific Evaluation Benchmarks for Each Language}
\label{sec:appendix-benchmarks}

\begin{table*}[!t]
    \centering
    \small
    \renewcommand{\arraystretch}{1.3} % Adjust row spacing for better readability
    \resizebox{\textwidth}{!}{%
    \begin{tabular}{m{3cm}|p{14cm}}  % m{} for vertical centering
        \toprule
        \multicolumn{1}{c|}{\textbf{Language}} & \multicolumn{1}{c}{\textbf{Evaluation Datasets}} \\
        \midrule
        \centering Arabic & ARC-C, Hellaswag~\citep{lai2023okapi}, 
        XNLI~\citep{conneau2018xnli}, XStoryCloze~\citep{lin2021few} \\
        \hline
        \centering English & ARC-E, ARC-C~\citep{clark2018think}, 
        Hellaswag~\citep{zellers2019hellaswag}, PAWS-X~\citep{yang-etal-2019-paws}, 
        PIQA~\citep{Bisk2020}, SciQ~\citep{SciQ}, 
        TruthfulQA~\citep{lin2021truthfulqa}, XNLI~\citep{conneau2018xnli}, 
        XStoryCloze~\citep{lin2021few} \\
        \hline
        \centering French & ARC-C, Hellaswag~\citep{lai2023okapi}, 
        PAWS-X~\citep{yang-etal-2019-paws}, XNLI~\citep{conneau2018xnli}, 
        XWinograd~\citep{tikhonov2021s} \\
        \hline
        \centering German & ARC-C, Hellaswag~\citep{lai2023okapi}, 
        PAWS-X~\citep{yang-etal-2019-paws}, XNLI~\citep{conneau2018xnli} \\
        \hline
        \centering Indonesian & ARC-C, Hellaswag~\citep{lai2023okapi}, 
        XCOPA~\citep{ponti2020xcopa}, XStoryCloze~\citep{lin2021few} \\
        \hline
        \centering Italian & ARC-C, Hellaswag~\citep{lai2023okapi}, 
        XCOPA~\citep{ponti2020xcopa} \\
        \hline
        \centering Russian & ARC-C, Hellaswag~\citep{lai2023okapi}, 
        XNLI~\citep{conneau2018xnli}, XStoryCloze~\citep{lin2021few}, 
        XWinograd~\citep{tikhonov2021s} \\
        \hline
        \centering Spanish & ARC-C, Hellaswag~\citep{lai2023okapi}, 
        PAWS-X~\citep{yang-etal-2019-paws}, XNLI~\citep{conneau2018xnli}, 
        XStoryCloze~\citep{lin2021few} \\
        \hline
        \centering Swahili & ARC-C, TruthfulQA~\citep{bayes2024uhura}, 
        XCOPA~\citep{ponti2020xcopa}, XNLI~\citep{conneau2018xnli}, 
        XStoryCloze~\citep{lin2021few} \\
        \hline
        \centering Welsh & ARC-E, ARC-C, 
        PIQA, TruthfulQA, and XNLI from \textit{BritEval} \\
        \bottomrule
    \end{tabular}%
    }
    \caption{Specific evaluation benchmarks for each language.}
    \label{tab:specific_benchmarks}
\end{table*}

Specific evaluation benchmarks for each of the 10 languages are shown in Table~\ref{tab:specific_benchmarks}.

\section{An Overview of Baseline Models}
\label{sec:appendix-baseline}
\begin{table*}[!t]
\small
\centering
\resizebox{1.0\textwidth}{!}{%
\begin{tabular}{lcccccc}
\toprule
\multirow{2}{*}{Model} & \multirow{2}{*}{\# Param.} & \multirow{2}{*}{Corpus} & Corpus & Training & Data & \multirow{2}{*}{Languages} \\
& & & Size & Tokens & Avail.& \\
\midrule
\multicolumn{7}{l}{\textit{Monolingual LLMs}} \\
\href{https://huggingface.co/TinyLlama/TinyLlama_v1.1}{TinyLlama} & 1.1B & {\makecell{SlimPajama~\citep{cerebras2023slimpajama} and \\ StarCoder training data~\citep{li2023starcoder}}} & 1T & 3T & \greencheck & Primarily English \\
\href{https://huggingface.co/EleutherAI/pythia-1.4b}{Pythia} & 1.4B & The Pile~\citep{gao2020pile} & 207B & 300B & \greencheck & Primarily English \\
\midrule
\multicolumn{7}{l}{\textit{Multilingual LLMs}} \\
\href{https://huggingface.co/ai-forever/mGPT}{mGPT} & 1.3B & mC4,Wiki & 488B & 440B & \redcross & {\makecell{61 languages}}\\
\href{https://huggingface.co/bigscience/bloomz-1b1}{BLOOM} & 1.1B & {\makecell{BigScience Catalogue, Common Crawl, Github Code, \\ and OSCAR~\citep{OrtizSuarezSagotRomary2019}}} & 350B & 366B & \redcross & {\makecell{46 langauges}} \\
\href{https://huggingface.co/meta-llama/Llama-3.2-1B}{Llama3.2} & 1.3B & {\makecell{Web data, Code, and Math}} &- & 9T & \redcross & {\makecell{At least 8 languages}}\\
\href{https://huggingface.co/Qwen/Qwen2-1.5B}{Qwen2} & 1.5B &- & - & 7T & \redcross & {\makecell{At least 30 languages}} \\
\href{https://huggingface.co/Qwen/Qwen2.5-1.5B}{Qwen2.5} & 1.5B &- & - & 18T & \redcross & {\makecell{At least 30 languages}} \\
\href{https://huggingface.co/google/gemma-2b}{Gemma} & 2.6B & Web data, Code, and Science Articles & - & 2T & \redcross & {\makecell{-}} \\
\midrule
\multicolumn{7}{l}{\textit{Language-specific LLMs}} \\

\href{https://huggingface.co/castorini/afriteva_v2_large}{afriteva\_v2\_large} & 1B & Wura~\citep{oladipo2023better} & 30 million & 136B & \greencheck & 20 African languages \\
\href{https://huggingface.co/britllm/britllm-3b-v0.1}{BritLLM} & 3B & {\makecell{SlimPajama~\citep{cerebras2023slimpajama}, \\ QA and MC Synthetic Data, Wiki, NLLB}} & 668B & - & \redcross & {\makecell{5 British languages}} \\

\href{https://huggingface.co/croissantllm/CroissantLLMBase}{CroissantLLM} & 1.3B & 
Croissant~\citep{faysse2024croissantllm} & 1T & 3T& \greencheck & English, French \\
\href{https://huggingface.co/utter-project/EuroLLM-1.7B}{EuroLLM} & 1.7B & {\makecell{mC4, Parallel Data, Code/Math, Wiki, \\ArXiv, Books, Apollo, Annealing Data}} & - & 4T & \redcross & {\makecell{35 languages}} \\

\href{https://huggingface.co/inceptionai/jais-family-1p3b}{Jais-family-1p3b} & 1.3B & {\makecell{Jais Model Family training data\\~\citep{sengupta2023jais}}} & 395B & 1.6T & \redcross & {\makecell{Arabic, English}} \\

\href{https://huggingface.co/sail/Sailor-1.8B}{Sailor} & 1.8B & {\makecell{CC100~\citep{wenzek-etal-2020-ccnet}, \\ MADLAD-400~\citep{kudugunta2024madlad}, \\ OpenSubtitles, and Wiki}} & 395B & 400B & \redcross & {\makecell{English, Chinese, and \\ 5 South-East Asian languages}} \\

\href{https://huggingface.co/sail/Sailor2-1B}{Sailor2} & 1B & {\makecell{CC100~\citep{wenzek-etal-2020-ccnet}, \\ MADLAD-400~\citep{kudugunta2024madlad}, \\ OpenSubtitles, and Wiki}} & - & 500B & \redcross & {\makecell{15 languages}} \\

\midrule
\themodel{} (Ours) & 1.3B & \thedata{} & 1.7T & 1.5T & \greencheck & 10 languages \\
\bottomrule
\end{tabular}
}
\caption{Overview of pretraining data among LLMs.}
\label{tab:model_comparison}
\end{table*}
An overview of baseline models are shown in Table~\ref{tab:model_comparison}.

\section{Translation Data Generation from the Mistral-7B-Instruct LLM}
\label{sec:appendix-cuatrollm}
We employ \textit{Mistral-7B-Instruct-v0.1}\footnote{\href{https://hf.co/mistralai/Mistral-7B-Instruct-v0.1}{\texttt{hf.co/mistralai/Mistral-7B-Instruct-v0.1}}} as our translation model. However, its efficacy when prompted for document-level translation, particularly with long-context English source documents, has not yet been verified. A recent related work by \citet{maini2024rephrasing} has empirically demonstrated that prompting an LLM to rephrase more than $300$ tokens could lead to information loss when rephrasing web data. 

\begin{figure*}[!t]
\includegraphics[width=1.0\textwidth]{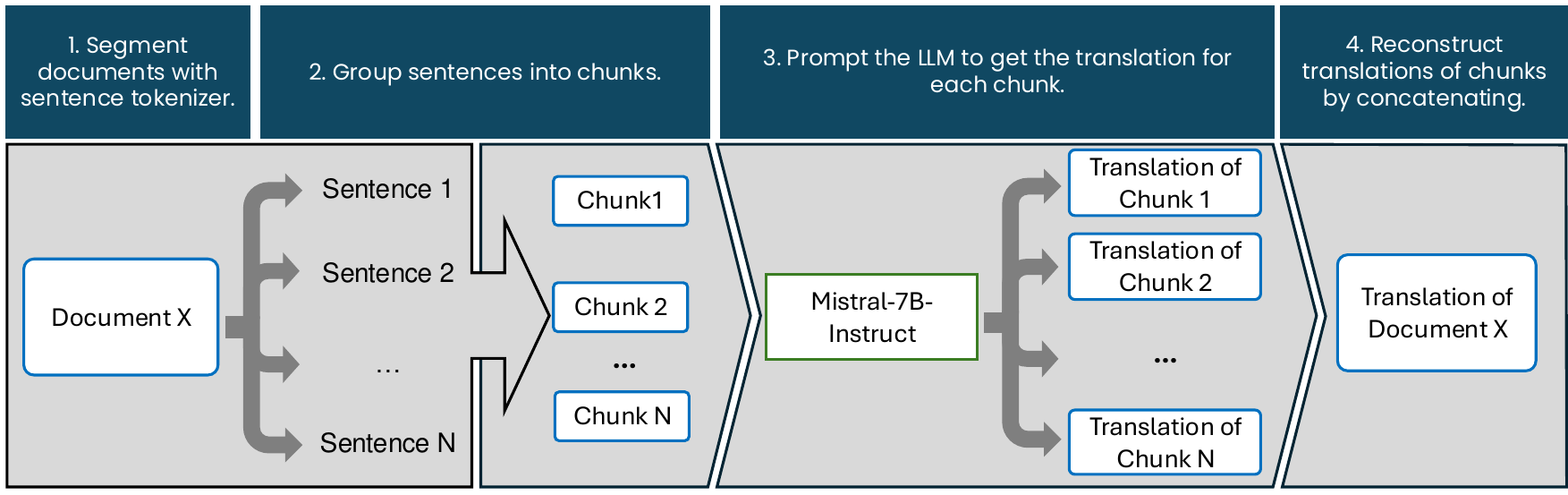}
\caption{Step-by-step illustration of the translation pipeline with the Mistral-7B-Instruct model.}
\label{fig:mistral_pip}
\end{figure*}

Following their setup, we first segment the English source documents from the sample-$100$BT subset of \textit{FineWeb-Edu} into shorter pieces, prompt Mistral to translate these segments sequentially, and subsequently reconstruct the whole translated document by concatenating the translated segments.The detailed translation pipeline is shown in Figure~\ref{fig:mistral_pip}.

\begin{figure}[!t]
\includegraphics[width=1.0\columnwidth]{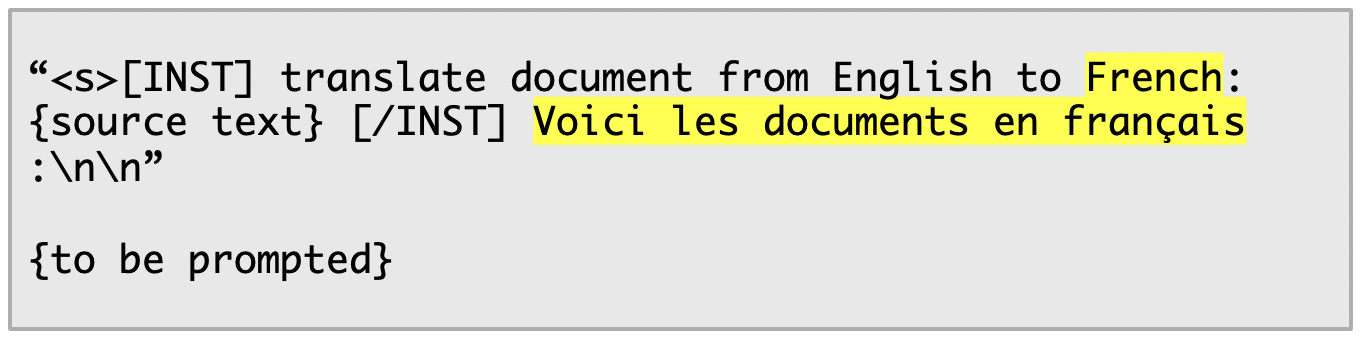}
\caption{Chat template used for prompting \textit{Mistral-7B-Instruct-v0.1} for English-French translation.}
\label{fig:prompt_template}
\end{figure}

Adhering to the instruction format specified\footnote{\href{https://hf.co/mistralai/Mistral-7B-Instruct-v0.1}{\texttt{hf.co/mistralai/Mistral-7B-Instruct-v0.1}}} for \textit{Mistral-7B-Instruct}, the chat template employed to prompt Mistral model for translation~(using English-French as an example) is illustrated in Figure~\ref{fig:prompt_template}\footnote{The highlighted portions in the template are adjusted according to the target language.}. 
% When prompting for the translation of each segment, we impose a maximum generation length of $384$ tokens. 
To maintain translation integrity, any sentence not fully translated to a terminal punctuation is omitted, based on the NLTK sentence tokenizer~\citep{bird2009natural}.

\begin{table}[!t]
\small
\centering
\begin{tabular}{l|cc}
\toprule
Language & \multicolumn{1}{c}{Tokens (B)} & \multicolumn{1}{c}{Avg. Doc Length (tokens)} \\
\midrule
English  & 63.41 & 1,171.48 \\
French   & 76.25 & 1,408.74 \\
German   & 73.91 & 1,365.41 \\
Spanish  & 72.93 & 1,383.25 \\
\midrule
Total &  286.50 & 1,331.89 \\
\bottomrule
\end{tabular}
\caption{Statistics of Translation Data generated with the Mistral-7B-Model, measured in Llama2 tokenizer.}
\label{tab:mistral_data_statistics}
\end{table}
We translate English documents from \textit{FineWeb-Edu} \citep{lozhkov2024fineweb-edu} into three major European languages: French, German, and Spanish via prompting the Mistral-7B-Intruct model. 
% We call this multiway parallel translated dataset as \cuadata. 
To optimize memory efficiency and accelerate the inference process of \textit{Mistral-7B-Instruct-v0.1}, we employ \textit{vLLM} \citep{kwon2023efficient}, a library specifically designed for efficient large language model inference and serving. Using this setup, we translate approximately $54$ million English documents (a subset of sample-$100$B of \textit{FineWeb-Edu}) into the three target languages by prompting \textit{Mistral-7B-Instruct-v0.1}. Table~\ref{tab:mistral_data_statistics} presents the statistics of the original English data and the translated French, German, and Spanish. Leveraging \textit{vLLM}'s efficiency, we estimate the total computational cost to be approximately $6.03 \times 10^{22}$ FLOPs.

\begin{table}[!t]
\scriptsize
\centering
\resizebox{\columnwidth}{!}{%
\begin{tabular}{l|cccc|c}
\toprule
 & en & fr & de & es & \textbf{Avg.} \\
\hline
\multicolumn{6}{@{}l@{}}{\textit{\textbf{English LLMs}}} \\
Pythia (1.4B) & 54.63 & 40.14 & 36.23 & 39.91 & 42.73 \\
TinyLlama (1.1B) & 57.18 & 44.13 & 37.06 & 42.38 & 45.19 \\
\hline
\multicolumn{6}{@{}l@{}}{\textit{\textbf{Multilingual LLMs}}} \\
mGPT (1.3B) & 45.38 & 40.27 & 34.54 & 38.88 & 39.77 \\
BLOOM (1.1B) & 50.94 & 41.80 & 34.62 & 42.69 & 42.51 \\
Llama3.2 (1.3B) & 58.16 & 44.10 & 39.73 & 44.15 & 46.54 \\
Qwen2 (1.5B) & 61.07 & 47.40 & 40.79 & 47.40 & 49.17 \\
Qwen2.5 (1.5B) & 62.93 & 48.69 & 40.49 & 48.41 & 50.13 \\
Gemma (2.6B) & 62.39 & 49.78 & 44.27 & 49.35 & 51.45 \\
\hline
\multicolumn{6}{@{}l@{}}{\textit{\textbf{Language-Specific LLMs}}} \\
CroissantLLM (1.3B) & 53.31 & 45.73 & - & - & - \\
EuroLLM (1.7B) & 57.98 & 47.85 & 42.07 & 47.07 & 48.74 \\
\hline
\multicolumn{6}{@{}l@{}}{\textit{\textbf{Ours}}} \\
CuatroLLM (1.3B) & 56.12 & 45.01 & 40.42 & 45.04 & 46.65 \\
TransWebLLM-4 (1.3B) & 55.32 & 46.67 & 41.45 & 45.19 & 47.16 \\
\bottomrule
\end{tabular}}
\caption{Performance comparison between \cuatrollm{} trained on LLM-translated data and \themodel{}-4 across four selected languages. 
% Scores represent model accuracy, with higher values indicating better performance. The last column reports the average accuracy across all four languages.
}
\label{tab:main_mistral_nllb_compare}
\end{table}

Table~\ref{tab:main_mistral_nllb_compare} compares the performance of the model trained on Mistral-generated translation data (\cuatrollm{}) with the model trained on NLLB-generated data (\themodel{}-4) across English, German, French, and Spanish benchmarks.

\section{Sampling General Web Data from \textit{RedPajama-v2} }
\label{sec:appendix-rpv2-sampling}
We use the English, French, German, Italian, and Spanish subsets of the \textit{RedPajama-v2}~(RPv2)~\cite{NEURIPS2024_d3449733} as web data. Given that web data is inherently noisy, we make further use of the quality signals provided for RPv2 and filter each subset down to a smaller, high-quality subset. Specifically, we use the six most recent dumps from 2022 and 2023 and apply quality filtering using the Gopher rules~\cite{rae2021scaling}. Additionally, web data often contains near duplicates, stemming from boilerplate text, ads, and other computer-generated text that only differs by a few words, and removing these has been shown to positively affect training efficiency and reduce the amount of memorization~\cite{lee2021deduplicating}. We therefore adopt the MinHash algorithm with locality-sensitive hashing~\cite{broder1997resemblance} to perform near-deduplication. We identify documents as near duplicates if their Jaccard similarity is greater than $0.8$ and use $128$ hash functions.

\section{Evaluation for Impact of Special Data}
\label{sec:appendix-special-data}
The evaluation results of \themodelcool{} on understanding and reasoning, French linguistic proficiency, and reasoning for Indonesian local culture are presented in Table~\ref{tab:all_understanding},~\ref{tab:french_ling_final}, and~\ref{tab:local_culture_reasoning_final}, respectively.

\begin{table*}[!t]
\centering
\small
\setlength{\tabcolsep}{4pt} 
\resizebox{1.0\textwidth}{!}{%
\begin{tabular}{l|ccccccc|c|cc|ccc}
\toprule
& \multicolumn{7}{c|}{\textit{High}} & \textit{Medium} & \multicolumn{2}{c|}{\textit{Low}} & \multicolumn{3}{c}{\textbf{Average}} \\
 & ar & en & fr & de & it & ru & es & id & sw & cy & All & Non-eng & High \\
\hline
\multicolumn{14}{@{}l@{}}{\textit{\textbf{English LLMs}}} \\
Pythia (1.4B) & 33.21 &  54.63 & 40.14 & 36.23 & 35.14 & 38.32 & 39.91 & 36.83 & 36.90 & 31.57  & 38.29 & 36.47 & 39.65 \\
TinyLlama (1.1B) & 32.86 & 57.18 & 44.13 & 37.06 & 36.79 & 40.97 & 42.38 & 36.13 & 36.69 & 31.53  & 39.57 & 37.62 & 41.62 \\
\hline
\multicolumn{14}{@{}l@{}}{\textit{\textbf{Multilingual LLMs}}} \\
mGPT (1.3B) & 32.90 & 45.38 & 40.27 & 34.54 & 34.89 & 40.53 & 38.88 & 39.47 & 38.98 & 31.14 & 37.70 & 36.84 & 38.20 \\
BLOOM (1.1B) & 34.97 & 50.94 & 41.80 & 34.62 & 33.69 & 37.56 & 42.69 & 43.23 & 37.09 & 31.57 & 38.82 & 37.47 & 39.47 \\
Llama3.2 (1.3B) & 34.78 & 58.16 & 44.10 & 39.73 & 40.93 & 45.38 & 44.15 & 44.67 & 38.30 & 31.84 & 42.20 & 40.43 & 43.89 \\
Qwen2 (1.5B) & 35.61 & \underline{61.07} & 47.40 & 40.79 & 42.61 & \underline{47.95} & \underline{47.40} & 45.93 & 38.89 & 32.23 & 43.99 & 42.09 & 46.12 \\
Qwen2.5 (1.5B) & 37.35 & \underline{\cellcolor{yellow!30}62.93} & \underline{48.69} & 40.49 & 43.10 & \underline{47.01} & \underline{48.41} & 46.17 & 37.98 & 31.78 & 44.39 & 42.33 & 46.85 \\
Gemma (2.6B) & 37.26 & \underline{62.39} & \underline{\cellcolor{yellow!30}49.78} & \underline{\cellcolor{yellow!30}44.27} & 44.57 & \underline{\cellcolor{yellow!30}48.60} & \underline{\cellcolor{yellow!30}49.35} & 48.27 & 40.18 & 32.28 & \underline{45.70} & \underline{43.84} & \underline{\cellcolor{yellow!30}48.03} \\
\hline
\multicolumn{14}{@{}l@{}}{\textit{\textbf{Language-Specific LLMs}}} \\
AfriTeVa (1B) & & 37.70 & & & & & & & 40.60 & & &\\
BritLLM (3B) & & 60.45 & & & & & & & & 37.07 & & &\\
CroissantLLM (1.3B) & & 53.31 & 45.73 & & & & & & & & &\\
EuroLLM (1.7B) & 38.88 & 57.98 & 47.85 & \underline{42.07} & \underline{47.56} & 46.71 & 47.07 & &  &  &  & & \underline{46.87}\\
% GPT-fr (1B) & & 0.3887 & 0.3736 & & & & & & & & & &\\
Jais-family-1p3b (1.3B) & \underline{39.97} & 56.28 & & & & & & & & & & &\\
Sailor (1.8B) & & 55.40 & & & & & & 48.45 & & & & & \\
Sailor2 (1B) & & 54.61 & & & & & & \underline{49.44} & & & & & \\
\hline
\multicolumn{14}{@{}l@{}}{\textit{\textbf{Ours}}} \\
TransWebLLM (1.3B) & 39.30 & 56.30 & 46.11 & 41.01 & 45.75 & 46.38 & 45.27 & 47.54 & \underline{44.44} & \underline{38.69} & 45.08 & 43.83 & 45.73 \\
TransWebLLM-web (1.3B) & \underline{39.82} & 56.18 & 46.83 & 41.92 & \underline{47.01} & 46.22 & 46.92 & \underline{49.75} & \underline{44.40} & \underline{40.09} & \underline{45.91} & \underline{44.77} & 46.41 \\
TransWebLLM-cool (1.3B) & \underline{\cellcolor{yellow!30}40.22} & 57.76 & \underline{48.38} & \underline{43.04} & \underline{\cellcolor{yellow!30}48.64} & 46.72 & 47.13 & \underline{\cellcolor{yellow!30}50.67} & \underline{\cellcolor{yellow!30}44.48} & \underline{\cellcolor{yellow!30}40.41} & \underline{\cellcolor{yellow!30}46.75} & \underline{\cellcolor{yellow!30}45.52} & \underline{47.41} \\
$\Delta$ (Cool - Base) & +0.92 & +1.46 & +2.27 & +2.03 & +2.89 & +0.34 & +1.86 & +3.13 & +0.04 & +1.72 & +1.67 & +1.69 & +1.68 \\
$\Delta$ (Cool - Web) & +0.40 & +1.58 & +1.55 & +1.12 & +1.63 & +0.50 & +0.21 & +0.92 & +0.08 & +0.32 & +0.84 & +0.75 & +1.00 \\
\bottomrule
\end{tabular}}
\caption{LLM performance across ten languages, categorized by resource availability and measured in accuracy. The last three columns report average results for all languages (All), non-English languages (Non-Eng), and high-resource languages (High). The top three models are underlined and the best score for each language is highlighted.}
\label{tab:all_understanding}
\end{table*}

\begin{table}[!t]
\centering
\small
\resizebox{0.40\textwidth}{!}{%
\begin{tabular}{l|cc|c}
\toprule
\textbf{Model} & \textit{\textbf{fr-grammar}} & \textit{\textbf{fr-vocab}} & \textbf{Avg.} \\
\midrule
\multicolumn{4}{@{}l@{}}{\textit{\textbf{Baselines}}} \\
mGPT & 73.95 & 70.59 & 72.27 \\
BLOOM & 79.83 & 74.79 & 77.31 \\
Llama3.2 & 76.47 & 75.63 & 76.05 \\
EuroLLM$^{*}$ & 79.83 & 78.99 & 79.41 \\
% Qwen2 & 71.43 & 73.95 & 72.69 \\
Qwen2.5 & 71.43 & 73.95 & 72.69 \\
Gemma & 73.11 & 72.27 & 72.69 \\
CroissantLLM$^{*}$ & 79.83 & 78.15 & 78.99 \\
\hline
\multicolumn{4}{@{}l@{}}{\textit{\textbf{Ours}}} \\
TransWebLLM & 67.23 & 63.03 & 65.13 \\
TransWebLLM-web & 73.11 & 76.47 & 74.79 \\
TransWebLLM-cool & 78.15 & 73.95 & 76.05 \\
\bottomrule
\end{tabular}}
\caption{French grammar and vocabulary proficiency evaluation for \themodelcool{}, measured in accuracy. The last column reports the task average score. Models with $^{*}$ denote regional models trained with support for French.}
\label{tab:french_ling_final}
\end{table}
\begin{table}[!t]
\centering
\small
\resizebox{0.40\textwidth}{!}{%
\begin{tabular}{l|cc|c}
\toprule
\textbf{Model} & \textbf{\textit{colloquial}} & \textbf{\textit{standard}} & \textbf{Avg.} \\
\midrule
\multicolumn{4}{@{}l@{}}{\textit{\textbf{Baselines}}} \\
mGPT & 54.56 & 53.49 & 54.03 \\
BLOOM & 55.10 & 54.38 & 54.74 \\
Llama3.2 & 52.42 & 52.59 & 52.51 \\
Qwen2.5 & 52.59 & 54.03 & 53.31 \\
Gemma & 54.03 & 55.99 & 55.01 \\
Sailor$^{*}$ & 57.60 & 65.47 & 61.54 \\
Sailor2$^{*}$ & 58.86 & 66.37 & 62.62 \\
\hline
\multicolumn{4}{@{}l@{}}{\textit{\textbf{Ours}}} \\
TransWebLLM & 48.12 & 49.55 & 48.84 \\
TransWebLLM-web & 55.46 & 59.75 & 57.61 \\
TransWebLLM-cool & 55.99 & 61.90 & 58.95 \\
\bottomrule
\end{tabular}}
\caption{COPAL-ID evaluation for \themodelcool{}, measured in accuracy. The last column reports task average score. Models with $^{*}$ denote regional models trained with support for Indonesian.}
\label{tab:local_culture_reasoning_final}
\end{table}

\section{Detailed Results for Understanding and Reasoning Benchmarks}
Tables~\ref{tab:ar_full} to~\ref{tab:cy_full} present detailed benchmark results, as outlined in Section~\ref{sec:benchmarks}, for each language. The corresponding averaged scores are presented in Tables~\ref{tab:main_result},~\ref{tab:main_real}, and~\ref{tab:all_understanding}, accordingly.

\begin{table}[h]
    \centering
    \renewcommand{\arraystretch}{1.2}
    \resizebox{1.0\columnwidth}{!}{
    \begin{tabular}{l|cccc|c}
        \toprule
        Model & ARC-C & Hellaswag & XNLI & XStoryCloze & Avg. \\
        \midrule
        Pythia & 21.04 & 27.16 & 36.31 & 48.31 & 33.21 \\
        TinyLlama & 20.79 & 26.87 & 35.26 & 48.51 & 32.86 \\
        mGPT & 20.44 & 26.02 & 35.02 & 50.10 & 32.90 \\
        BLOOM & 21.81 & 29.69 & 35.90 & 52.48 & 34.97 \\
        Llama3.2 & 22.58 & 30.47 & 34.86 & 51.22 & 34.78 \\
        EuroLLM & 25.58 & 33.93 & 37.63 & 58.37 & 38.88 \\
        Qwen2 & 23.95 & 31.26 & 34.22 & 53.01 & 35.61 \\
        Qwen2.5 & 27.29 & 32.14 & 34.82 & 55.13 & 37.35 \\
        Gemma & 26.69 & 32.32 & 35.54 & 54.47 & 37.26 \\
        jais-family-1p3b & 28.14 & 35.51 & 36.47 & 59.76 & 39.97 \\
        \midrule
        TransWebLLM & 29.43 & 34.26 & 36.39 & 57.11 & 39.30 \\
        TransWebLLM-web & 28.57 & 35.95 & 36.51 & 58.24 & 39.82 \\
        TransWebLLM-cool & 30.28 & 35.90 & 35.78 & 58.90 & 40.22 \\
        \bottomrule
    \end{tabular}}
    \caption{Detailed Arabic Benchmark Results.}
    \label{tab:ar_full}
\end{table}

\begin{table}[h]
    \centering
    \renewcommand{\arraystretch}{1.2}
    \resizebox{1.0\columnwidth}{!}{
    \begin{tabular}{l|ccccccccc|c}
        \toprule
        Model & ARC-C & ARC-E & Hellaswag & PAWS & PIQA & SciQ & TruthfulQA & XNLI & XStoryCloze & Avg. \\
        \midrule
        Pythia & 27.99 & 64.23 & 40.49 & 56.55 & 71.06 & 91.80 & 22.77 & 48.59 & 68.17 & 54.63 \\
        TinyLlama & 33.79 & 68.01 & 46.52 & 57.40 & 74.16 & 93.60 & 22.28 & 46.95 & 71.87 & 57.18 \\
        mGPT & 21.93 & 48.99 & 30.65 & 53.90 & 64.53 & 63.10 & 23.26 & 41.81 & 60.23 & 45.38 \\
        BLOOM & 24.91 & 54.59 & 34.69 & 52.05 & 67.85 & 89.30 & 25.58 & 47.39 & 62.14 & 50.94 \\
        EuroLLM & 36.95 & 71.51 & 44.77 & 55.90 & 73.56 & 94.80 & 23.62 & 48.92 & 71.81 & 57.98 \\
        Llama3.2 & 34.81 & 69.11 & 48.29 & 55.90 & 75.57 & 95.20 & 23.38 & 48.55 & 72.67 & 58.16 \\
        Qwen2 & 40.02 & 72.81 & 49.17 & 63.30 & 75.84 & 96.00 & 29.01 & 48.47 & 75.05 & 61.07 \\
        Qwen2.5 & 48.55 & 80.47 & 49.98 & 58.30 & 77.04 & 96.80 & 29.25 & 51.20 & 74.78 & 62.93 \\
        Gemma & 47.61 & 77.31 & 52.88 & 61.70 & 76.88 & 96.80 & 21.91 & 49.56 & 76.84 & 62.39 \\
        Afriteva-v2-large & 20.99 & 30.93 & 26.47 & 50.90 & 55.88 & 43.70 & 25.21 & 36.14 & 49.11 & 37.70 \\
        BritLLM & 38.05 & 72.81 & 51.15 & 60.65 & 75.84 & 96.10 & 24.36 & 49.40 & 75.71 & 60.45 \\
        CroissantLLM & 26.37 & 62.58 & 40.88 & 52.05 & 72.69 & 92.70 & 23.62 & 42.89 & 66.05 & 53.31 \\
        Jais-family-1p3b & 29.52 & 64.90 & 42.59 & 60.10 & 72.58 & 93.90 & 25.46 & 48.84 & 68.63 & 56.28 \\
        Sailor & 29.52 & 64.02 & 42.79 & 56.70 & 72.96 & 92.80 & 22.03 & 47.95 & 69.82 & 55.40 \\
        Sailor2 & 30.03 & 64.18 & 40.00 & 56.55 & 70.67 & 94.50 & 22.89 & 45.74 & 66.91 & 54.61 \\
        \midrule
        TransWebLLM & 36.86 & 71.51 & 40.65 & 56.70 & 71.06 & 93.00 & 22.89 & 48.15 & 65.92 & 56.30 \\
        TransWebLLM-web & 36.77 & 71.84 & 41.10 & 55.90 & 70.13 & 93.30 & 21.66 & 47.31 & 67.57 & 56.18 \\
        TransWebLLM-cool & 38.65 & 72.98 & 42.13 & 59.50 & 71.71 & 93.80 & 25.58 & 47.83 & 67.64 & 57.76 \\
        \bottomrule
    \end{tabular}}
    \caption{Detailed English Benchmark Results.}
    \label{tab:en_full}
\end{table}

\begin{table}[h]
    \centering
    \renewcommand{\arraystretch}{1.2}
    \resizebox{1.0\columnwidth}{!}{
    \begin{tabular}{l|ccccc|c}
        \toprule
        Model & ARC-C & Hellaswag & PAWS & XNLI & XWinograd & Avg. \\
        \midrule
        Pythia & 20.62 & 29.60 & 52.45 & 43.82 & 54.22 & 40.14 \\
        TinyLlama & 24.64 & 32.68 & 53.40 & 42.45 & 67.47 & 44.13 \\
        mGPT & 20.87 & 27.13 & 53.75 & 40.56 & 59.04 & 40.27 \\
        BLOOM & 23.18 & 33.73 & 52.10 & 45.78 & 54.22 & 41.80 \\
        Llama3.2 & 27.29 & 36.04 & 50.80 & 43.73 & 62.65 & 44.10 \\
        EuroLLM & 32.25 & 40.25 & 53.05 & 45.02 & 68.67 & 47.85 \\
        Qwen2 & 29.34 & 38.76 & 56.95 & 44.50 & 67.47 & 47.40 \\
        Qwen2.5 & 33.45 & 38.05 & 57.55 & 44.54 & 69.88 & 48.69 \\
        Gemma & 35.41 & 39.81 & 59.95 & 47.47 & 66.27 & 49.78 \\
        CroissantLLM & 25.75 & 39.70 & 52.65 & 44.30 & 66.27 & 45.73 \\
        \midrule
        TransWebLLM & 35.16 & 38.98 & 53.45 & 45.14 & 57.83 & 46.11 \\
        TransWebLLM-web & 36.27 & 40.37 & 53.20 & 44.06 & 60.24 & 46.83 \\
        TransWebLLM-cool & 36.78 & 40.95 & 55.80 & 45.70 & 62.65 & 48.38 \\
        \bottomrule
    \end{tabular}}
    \caption{Detailed French Benchmark Results.}
    \label{tab:fr_full}
\end{table}
\begin{table}[h]
    \centering
    \renewcommand{\arraystretch}{1.2}
    \resizebox{1.0\columnwidth}{!}{
    \begin{tabular}{l|cccc|c}
        \toprule
        Model & ARC-C & Hellaswag & PAWS& XNLI & Avg. \\
        \midrule
        Pythia & 19.59 & 28.53 & 55.20 & 41.61 & 36.23 \\
        TinyLlama & 21.56 & 30.66 & 55.55 & 40.48 & 37.06 \\
        mGPT & 19.42 & 27.69 & 50.50 & 40.56 & 34.54 \\
        BLOOM & 20.19 & 27.18 & 53.30 & 37.79 & 34.62 \\
        Llama3.2 & 26.86 & 34.09 & 54.05 & 43.90 & 39.73 \\
        EuroLLM & 29.34 & 37.51 & 55.25 & 46.18 & 42.07 \\
        Qwen2 & 26.60 & 34.72 & 59.20 & 42.65 & 40.79 \\
        Qwen2.5 & 28.23 & 34.81 & 56.00 & 42.93 & 40.49 \\
        Gemma & 31.05 & 37.32 & 62.50 & 46.22 & 44.27 \\
        \midrule
        TransWebLLM & 32.59 & 36.07 & 51.60 & 43.78 & 41.01 \\
        TransWebLLM-web & 31.65 & 37.60 & 54.50 & 43.94 & 41.92 \\
        TransWebLLM-cool & 33.02 & 37.83 & 55.65 & 45.66 & 43.04 \\
        \bottomrule
    \end{tabular}}
    \caption{Detailed German Benchmark Results.}
    \label{tab:de_full}
\end{table}

\begin{table}[h]
    \centering
    \renewcommand{\arraystretch}{1.2}
    \resizebox{0.8\columnwidth}{!}{
    \begin{tabular}{l|ccc|c}
        \toprule
        Model & ARC-C& Hellaswag & XCOPA & Avg. \\
        \midrule
        Pythia & 21.30 & 29.12 & 55.00 & 35.14 \\
        TinyLlama & 23.01 & 31.17 & 56.20 & 36.79 \\
        mGPT & 19.33 & 27.34 & 58.00 & 34.89 \\
        BLOOM & 20.79 & 28.47 & 51.80 & 33.69 \\
        Llama3.2 & 27.46 & 34.94 & 60.40 & 40.93 \\
        EuroLLM & 33.53 & 39.56 & 69.60 & 47.56 \\
        Qwen2 & 28.23 & 35.81 & 63.80 & 42.61 \\
        Qwen2.5 & 31.74 & 35.57 & 62.00 & 43.10 \\
        Gemma & 31.82 & 37.49 & 64.40 & 44.57 \\
        \midrule
        TransWebLLM & 36.53 & 37.33 & 63.40 & 45.75 \\
        TransWebLLM-web & 36.01 & 39.01 & 66.00 & 47.01 \\
        TransWebLLM-cool & 37.13 & 39.38 & 69.40 & 48.64 \\
        \bottomrule
    \end{tabular}}
    \caption{Detailed Italian Benchmark Results.}
    \label{tab:it_full}
\end{table}

\begin{table}[h]
    \centering
    \renewcommand{\arraystretch}{1.2}
    \resizebox{1.0\columnwidth}{!}{
    \begin{tabular}{l|ccccc|c}
        \toprule
        Model & ARC-C & Hellaswag & XNLI & XStoryCloze & XWinograd & Avg. \\
        \midrule
        Pythia & 18.99 & 27.66 & 39.72 & 49.37 & 55.87 & 38.32 \\
        TinyLlama & 22.93 & 30.60 & 39.32 & 53.61 & 58.41 & 40.97 \\
        mGPT & 20.27 & 26.62 & 40.04 & 56.65 & 59.05 & 40.53 \\
        BLOOM & 19.85 & 27.52 & 38.07 & 48.05 & 54.29 & 37.56 \\
        Llama3.2 & 25.92 & 34.17 & 42.57 & 60.09 & 64.13 & 45.38 \\
        EuroLLM & 29.00 & 36.69 & 45.14 & 62.74 & 60.00 & 46.71 \\
        Qwen2 & 27.63 & 36.18 & 42.29 & 64.46 & 69.21 & 47.95 \\
        Qwen2.5 & 30.88 & 35.93 & 42.29 & 61.81 & 64.13 & 47.01 \\
        Gemma & 33.11 & 36.79 & 45.50 & 62.54 & 65.08 & 48.60 \\
        \midrule
        TransWebLLM & 32.85 & 35.63 & 41.16 & 58.77 & 63.49 & 46.38 \\
        TransWebLLM-web & 32.42 & 37.03 & 40.88 & 59.83 & 60.95 & 46.22 \\
        TransWebLLM-cool & 33.28 & 37.10 & 40.60 & 61.35 & 61.27 & 46.72 \\
        \bottomrule
    \end{tabular}}
    \caption{Detailed Russian Benchmark Results.}
    \label{tab:ru_full}
\end{table}
\begin{table}[h]
    \centering
    \renewcommand{\arraystretch}{1.2}
    \resizebox{1.0\columnwidth}{!}{
    \begin{tabular}{l|ccccc|c}
        \toprule
        Model & ARC-C & HellaSwag & PAWS & XNLI & XStoryCloze & Avg. \\
        \midrule
        Pythia & 22.22 & 30.28 & 52.00 & 41.45 & 53.61 & 39.91 \\
        TinyLlama& 24.10 & 33.38 & 54.40 & 42.77 & 57.25 & 42.38 \\
        mGPT & 20.51 & 28.43 & 50.40 & 39.88 & 55.20 & 38.88 \\
        BLOOM & 24.87 & 34.48 & 51.55 & 43.90 & 58.64 & 42.69 \\
        Llama3.2 & 28.63 & 37.08 & 50.70 & 42.73 & 61.61 & 44.15 \\
        EuroLLM & 33.08 & 41.08 & 52.50 & 44.38 & 64.33 & 47.07 \\
        Qwen2 & 30.43 & 39.03 & 59.35 & 43.13 & 65.06 & 47.40 \\
        Qwen2.5 & 35.56 & 39.56 & 59.25 & 43.49 & 64.20 & 48.41 \\
        Gemma & 35.98 & 41.69 & 59.35 & 44.10 & 65.65 & 49.35 \\
        \midrule
        TransWebLLM & 34.79 & 39.17 & 50.80 & 43.29 & 58.31 & 45.27 \\
        TransWebLLM-web & 36.07 & 40.72 & 54.00 & 43.25 & 60.56 & 46.92 \\
        TransWebLLM-cool & 35.81 & 40.94 & 54.60 & 43.37 & 60.95 & 47.13 \\
        \bottomrule
    \end{tabular}}
    \caption{Detailed Spanish Benchmark Results.}
    \label{tab:es_full}
\end{table}

\begin{table}[h]
    \centering
    \renewcommand{\arraystretch}{1.2}
    \resizebox{1.0\columnwidth}{!}{
    \begin{tabular}{l|cccc|c}
        \toprule
        Model & ARC-C & HellaSwag & XCOPA & XStoryCloze & Avg. \\
        \midrule
        Pythia & 17.01 & 27.78 & 53.60 & 48.91 & 36.83 \\
        TinyLlama & 15.90 & 27.44 & 52.20 & 48.97 & 36.13 \\
        mGPT & 18.97 & 27.08 & 58.20 & 53.61 & 39.47 \\
        BLOOM & 21.37 & 31.70 & 62.00 & 57.84 & 43.23 \\
        Llama3.2 & 23.93 & 34.13 & 61.40 & 59.23 & 44.67 \\
        Qwen2 & 26.07 & 34.34 & 62.60 & 60.69 & 45.93 \\
        Qwen2.5 & 27.26 & 34.66 & 62.80 & 59.96 & 46.17 \\
        Gemma & 31.45 & 36.35 & 64.40 & 60.89 & 48.27 \\
        Sailor & 26.84 & 36.26 & 68.40 & 62.28 & 48.45 \\
        Sailor2 & 27.61 & 36.82 & 69.60 & 63.73 & 49.44 \\
        \midrule
        TransWebLLM & 34.87 & 37.05 & 60.60 & 57.64 & 47.54 \\
        TransWebLLM-web & 33.50 & 37.90 & 65.80 & 61.81 & 49.75 \\
        TransWebLLM-cool & 34.36 & 37.84 & 68.40 & 62.08 & 50.67 \\
        \bottomrule
    \end{tabular}}
    \caption{Detailed Indonesian Benchmark Results}
    \label{tab:id_full}
\end{table}

\begin{table}[h]
    \centering
    \renewcommand{\arraystretch}{1.2}
    \resizebox{1.0\columnwidth}{!}{
    \begin{tabular}{l|ccccc|c}
        \toprule
        Model & ARC-C & TruthfulQA & XCOPA & XNLI & XStoryCloze & Avg. \\
        \midrule
        Pythia & 22.81 & 24.54 & 53.60 & 34.42 & 49.11 & 36.90 \\
        TinyLlama & 22.00 & 24.78 & 52.60 & 34.30 & 49.77 & 36.69 \\
        mGPT & 27.70 & 24.41 & 56.00 & 35.50 & 51.29 & 38.98 \\
        BLOOM & 22.61 & 25.77 & 53.00 & 34.30 & 49.77 & 37.09 \\
        Llama3.2 & 28.31 & 24.16 & 52.60 & 34.46 & 51.95 & 38.30 \\
        Qwen2 & 26.88 & 29.49 & 53.60 & 34.70 & 49.77 & 38.89 \\
        Qwen2.5& 25.05 & 27.39 & 53.40 & 34.58 & 49.50 & 37.98 \\
        Gemma & 27.09 & 25.03 & 56.00 & 37.87 & 54.93 & 40.18 \\
        Afriteva\_v2\_large & 27.70 & 35.32 & 55.40 & 34.94 & 49.64 & 40.60 \\
        \midrule
        TransWebLLM & 27.70 & 33.95 & 62.00 & 41.57 & 56.98 & 44.44 \\
        TransWebLLM-web & 27.70 & 28.38 & 64.80 & 42.01 & 59.10 & 44.40 \\
        TransWebLLM-cool & 33.81 & 22.06 & 64.80 & 42.81 & 58.90 & 44.48 \\
        \bottomrule
    \end{tabular}}
    \caption{Detailed Swahili Benchmark Results.}
    \label{tab:sw_full}
\end{table}

\begin{table}[h]
    \centering
    \renewcommand{\arraystretch}{1.2}
    \resizebox{1.0\columnwidth}{!}{
    \begin{tabular}{l|ccccc|c}
        \toprule
        Model & ARC-C & ARC-E& PIQA & TruthfulQA& XNLI& Avg. \\
        \midrule
        Pythia & 17.68 & 25.87 & 51.88 & 27.60 & 34.84 & 31.57 \\
        TinyLlama & 18.37 & 26.42 & 52.05 & 28.00 & 32.83 & 31.53 \\
        mGPT & 17.51 & 25.83 & 53.40 & 24.80 & 34.14 & 31.14 \\
        BLOOM & 18.37 & 26.98 & 52.50 & 24.93 & 35.05 & 31.57 \\
        Llama3.2 & 18.20 & 26.34 & 53.85 & 25.87 & 34.92 & 31.84 \\
        Qwen2 & 18.88 & 25.70 & 52.89 & 28.80 & 34.88 & 32.23 \\
        Qwen2.5 & 18.37 & 26.30 & 51.54 & 27.07 & 35.62 & 31.78 \\
        Gemma & 18.54 & 27.91 & 53.00 & 27.07 & 34.88 & 32.28 \\
        BritLLM & 21.97 & 40.36 & 58.56 & 24.13 & 40.35 & 37.07 \\
        \midrule
        TransWebLLM & 26.87 & 43.37 & 56.54 & 27.07 & 39.61 & 38.69 \\
        TransWebLLM-web & 28.41 & 47.03 & 57.38 & 26.40 & 41.22 & 40.09 \\
        TransWebLLM-cool & 28.41 & 47.79 & 57.78 & 26.80 & 41.26 & 40.41 \\
        \bottomrule
    \end{tabular}}
    \caption{Detailed Welsh Benchmark Results.}
    \label{tab:cy_full}
\end{table}

\end{document}